\documentclass[preprint, 12pt]{elsarticle}

\usepackage{amssymb}
\usepackage{amsthm}

\usepackage{lineno,hyperref}
\modulolinenumbers[5]

\usepackage{color}
\usepackage[none]{hyphenat}
\usepackage{booktabs}
\usepackage{tabularx}
\usepackage{longtable} 
\usepackage{url}
\usepackage{multirow}
\usepackage{enumitem}
\usepackage{pgf}
\usepackage[utf8]{inputenc}
\usepackage[T1]{fontenc}

\usepackage{array}
\newcolumntype{P}[1]{>{\centering\arraybackslash}p{#1}}

\usepackage{pdflscape}
\usepackage{geometry}
\usepackage{hhline}

\newcommand{\transen}[1]{\footnote{English translation: {#1}}}

\usepackage[colorinlistoftodos]{todonotes}

\newtheorem{adib}{Example}

\biboptions{authoryear}

\journal{Engineering Applications of Artificial Intelligence}

\begin{document}

\begin{frontmatter}



\title{Real Time Monitoring of Social Media and Digital Press}


\author[elh]{Iñaki San Vicente*, Xabier Saralegi}
\author[ehu]{Rodrigo Agerri}

\address[elh]{Elhuyar Foundation}
\address[ehu]{IXA NLP Group, University of the Basque Country UPV/EHU}

\begin{abstract}
    Talaia is a platform for monitoring social media and digital press. A configurable crawler gathers content with respect to user defined domains or topics. Crawled data is processed by means of the EliXa Sentiment Analysis system. A Django powered interface provides data visualization for a user-based analysis of the data. This paper presents the architecture of the system and describes in detail its different components. To prove the validity of the approach, two real use cases are accounted for: one in the cultural domain and one in the political domain. Evaluation for the sentiment analysis task in both scenarios is also provided, showing the capacity for domain adaptation.
\end{abstract}

\begin{keyword}
Sentiment Analysis \sep Social Media Analysis \sep Crawling \sep Natural Language Processing \sep Digital Media Monitoring

\end{keyword}

\end{frontmatter}


\section{Introduction}
\label{isv2018:intro}

The Internet is a very rich source of user-generated information. As knowledge management technologies have evolved, many organizations have turned their eyes to such information, as a way of obtaining global feedback on their activities \citep{chen2012business}. Some studies \citep{oconnor_tweets_2010,ceron2015using} have pointed out that such systems could perform as well as traditional polling systems, but at a much lower cost. 

Talaia is a platform for monitoring the impact of topics specified by the user in social media and digital press. The process starts when the user configures the system to find information related to a domain or topic. Talaia provides real time information on the topic and graphic visualizations to help users interpreting the data. Such technology has various applications areas, such as:
\begin{itemize}
\item Monitoring events: Following public events in real time harvesting people's opinions and media news \citep{sutton2009social,yu2015world}.
\item Analyze citizen or electors voice: Tracking the opinions citizens convey with respect to public services or trends during electoral campaigns \citep{ceron2015using}.
\item Marketing and brand management: Measuring the impact of marketing campaigns in a digital environment \citep{ahmed2018measuring}.
\item Business Intelligence: Fast and efficient visualization of the information extracted from social media offers companies the possibility to analyse opinions about their products or services 	\citep{he2013social,mostafa2013more}. 
\item Security: Detection of social conflicts, crimes, and cyberbullying \citep{Xuetal2012sabullying,dadvar2013improving}.
\end{itemize}

\begin{figure*}[!ht]
	\centering
	\includegraphics[width=0.6\textwidth]{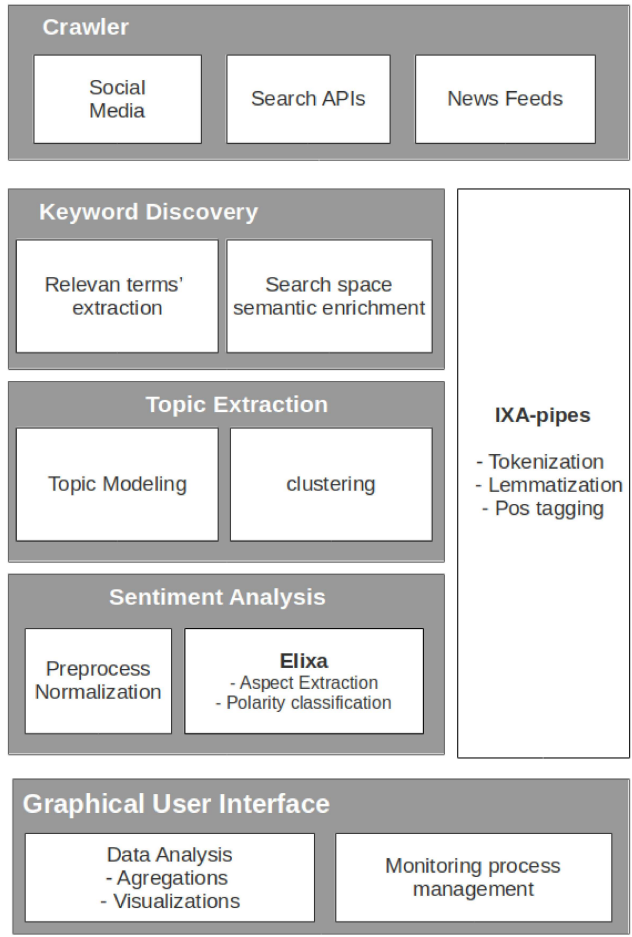}
	\caption{Diagram Talaia's components.}
	\label{isv2018:fig:arch}
\end{figure*}

Talaia consists of three main modules: (i) a crawler collecting the data; (ii) a data analysis module; and (iii) a Graphical User Interface (GUI) providing interpretation of the data analyzed. Figure \ref{isv2018:fig:arch} describes the architecture. Its main features are the following: 

\begin{itemize}
\item Monitoring and automatic analysis: Definition of the domain/topic by means of term taxonomies. Continuous monitoring of various mention sources including social media and digital press.

\item Multilingual extraction of mentions and opinions relevant to the topics monitored, by means of Natural Language Processing (NLP) techniques. 

\item Result exploration: Intuitive GUI to visualize and analyse the results. Advanced statistics and filters, such as per language results, impact of the topics or author statistics.

\item Control of the monitoring process through the user interface: update search terms or review and correct gathered mentions.

\end{itemize}

This paper focuses on those processes monitoring user satisfaction with respect to a topic, and that is why we pay special attention to the Sentiment Analysis (SA) module. Nevertheless, Talaia is capable of performing further data analysis tasks involving user profiling, in order to get the most out of the data. Specifically, geolocalization, user community identification and gender detection have been implemented. Section \ref{isv2018:da:profiling} provides more details.

The rest of the paper is organized as follows. Section \ref{isv2018:soa} discusses previous work, focusing on social media and SA. Both the academic and industrial points of view are taken into account. The third section describes in detail the modules composing Talaia. Section \ref{isv2018:success} presents two success cases where the platform has been used for monitoring different events. Section \ref{isv2018:eval} provides evaluation and results on the SA task for both scenarios. The last section draws some conclusions and future directions.

\section{Background}
\label{isv2018:soa}

\subsection{Social Media Analysis}
\label{isv2018:soa:SocialMedia}

Social media are becoming the primary environment for producing, spreading and consuming information. Enormous quantities of user generated content are produced constantly. Even traditional media spread their news and get a large amount of traffic trough social media. Monitoring events or topics in such an environment is however a challenging task. That is where data mining and Natural Language Processing (NLP) become essential. We have to be able to collect large scale data, but also to extract the relevant information. Tracking a topic over an extended time period means that the information flow grows and fades over time. Also a topic may evolve in terms of the vocabulary used, and thus ``topic detection and tracking'' (TDT) \citep{allan1998topic} techniques become relevant to maintain a successful monitoring.

Several systems have been proposed in the literature to explore events. Trend Miner \citep{preoctiuc2013temporal} extracts multilingual terms from social media, groups and visualizes them in temporal series. Social Sensor \citep{aiello2013sensing}
and Twitcident \citep{abel2012twitcident} may be the most similar systems to ours. The first one focuses on tracking topic or events predefined by the user. The second makes user defined searches related to crisis management. LRA\footnote{\url{https://www.lracrisistracker.com}} aims to discovering and tracking crisis situations based on crowdsourced information. ReDites \citep{osborne2014real} detects an tracks topics in a fully automated way.

 
Detecting terms that represent a domain or topic semantically has been traditionally addressed by statistical models such as Latent Dirichlet Association (LDA) \citep{blei2003latent}. Classical LDA models are applied over static document collections. In order to extract terms from dynamic collections, the most common approach is to follow a two step strategy \citep{shamma2011peaks} consisting of detecting emerging terms and grouping them in clusters thereby defining a domain.

\citet{nguyen2016hot} predict emerging terms by means of word co-occurrence distributional models, comparing the terms in an specific time window against the whole collection. \citet{abilhoa2014keyword} use a graph-based representation of the document collection. \citet{aiello2013sensing} propose $df$-$idf_t$ (Document Frequency - Inverse Document Frequency), a variation of $tf$-$idf$ that includes the temporal factor. \citet{kim2016exploring} combine neural networks and sequence labelling in order to extract relevant terms from conversations. \citet{miao2017cost} propose to reduce the cost of predicting emerging topics by finding a small group of representative users and predict the emerging topics from their social media activity.

There is also the problem of the scope of the event or topic to be tracked. An event might be tracked at global level (e.g. Football World Cup), but most events are local or regional at most. Two issues arise at this point. Firstly, how to restrict the data gathered to a specific region, and, secondly, how to cope with multilingual data. Some authors tackle the problem by automatically geolocating tweets while others try focus on user locations. See \citet{jurgens2015geolocation,zubiaga2017towards} for reviews of previous approaches. Our approach is to geolocate users rather than tweets, in order to construct a census of Twitter users in a region. We developed a SVM classifier similar to \citet{zubiaga2017towards} using follower and friend information as features.

\subsection{Sentiment Analysis}
\label{isv2018:soa:SA}


In the last years microblogging sites such as Twitter have attracted the attention of many researchers with diverse objectives such as stock market prediction \citep{bollen_twitter_2010,oliveira2017impact}, polling estimation \citep{oconnor_tweets_2010,ceron2015using} or analysis of crisis situations \citep{griffith2016analysis,shaikh2017tweet,ozturk2018sentiment}. The growing number of SA related shared tasks (e.g., SemEval Aspect based SA and Twitter SA shared tasks) or the commercial platforms for reputation management (see section \ref{isv2018:soa:indust}) are proof of the interest from both academic and market worlds.

The particularities of its language make it hard to analyze tweets. User mentions, hashtags, the growing presence of emojis, ungrammatical sentences, vocabulary variations and other phenomena pose a great challenge for traditional NLP tools \citep{foster_hardtoparse2011,liu_recognizing_2011}. \citet{brody_cool2011} deal with the word lengthening phenomenon, which is especially important for sentiment analysis because it usually expresses emphasis of the message. Hashtag decomposition (e.g.,\textit{\#GameOfThrones = `Game Of Thrones'}) \citep{BrunRoux_F14-2015,Belainine_W16-3915} or matching Out Of Vocabulary (OOV) forms and acronyms to their standard vocabulary forms (e.g., {\it `imo = in my opinion'}) \citep{han_lexical_2011,liu_broad-coverage_2012,alegria2014tweetnorm} are other addressed issues. International benchmarking initiatives such as the TweetNorm shared task\citep{tweetnorm_jlre_2015} or the WNUT\footnote{\url{http://noisy-text.github.io/2018/}} workshop series are proof of the interest to solve this task.

Once texts are normalized, sentiment analysis can be performed. Several ruled-based systems to polarity classification have been proposed \citep{hu_mining_2004,thelwall2017heart,taboada2011lexicon}. Nevertheless, we will focus on Machine Learning (ML) based approaches which are the most widespread. Support Vector Machines (SVM) and Logistic Regression algorithms have been the very popular for polarity classification as various international shared tasks \citep{roman_tass2014,semeval-2014_4,semeval-2014_9} show. Typical features of those systems include sentiment word/lemma ngram features, POS tags \citep{barbosa_robust_2010}, Sentiment Lexicons \citep{kouloumpis_twitter_2011}, emoticons \citep{oconnor_tweets_2010}, discourse information \citep{somasundaran_supervised_2009} or, more recently, word embeddings \citep{mikolov2013distributed} and clusters \citep{elixa-semeval2015}.

From 2015 onwards, the academic world has shifted to Deep Learning (DL) approaches, as \citet{Semeval2016-Twitter-task4} confirm. Long-Short Term Memory (LSTM) Recurrent Neural Networks (RNN)\citep{DaiLe2015NIPS,johnson2016supervised} and Convolutional Neural Networks (CNN) are the preferred choices. \citet{severyn2015} use a single layer CNN, first to construct word embeddings and then to train the classifier. \citet{Deriu2017} propose a two phase training method: first they train a neural network with large amounts of weakly supervised data collected from Twitter. The network is initialized with word embeddings learned by means of  word2vec \citep{mikolov2013distributed} from very large corpora collected from twitter. Second, the weights learned in the first step are transferred to a second neural network trained over the actual annotated data, to learn the final classifier. A two convolutional layer CNN is used for both training phases. A very similar approach is followed by \citet{Cliche2017} which achieves top results in SemEval \citep{Semeval2017-twitter-task4}. \citet{ulmfit2018} follow a similar three step approach with a more complex network topology obtaining state of the art results for various tasks, including sentiment analysis.

A common problem of supervised approaches, specially of DL, is the need of large amounts of labeled data for training. The common practice in the literature is to gather weakly supervised datasets following the emoticon heuristic\footnote{Collect tweets containing the ``:)'' emoticon and regard them as positive, and likewise for the ``:('' emoticon.} \citep{go_twitter_2009}. This is feasible for major languages, but it is a very difficult (if possible at all) and time costly task for non major languages such as Basque.

Our system is closest to \citet{barbosa_robust_2010} and \citet{kouloumpis_twitter_2011} because it combines polarity lexicons with machine learning for labelling sentiment of tweets. This strategy has proven to be a successful approach in previous shared tasks \citep{saralegi_tass_2012,mohammad_semeval_2013}.

\subsection{Industrial Solutions}
\label{isv2018:soa:indust}

We can find various commercial solutions in the market. We are particularly interested in systems that provide an integral solution of the monitoring process, leaving out tools that only approach specific phases of the surveillance process, or solutions that offer bare NLP processing chains which require further development to achieve a working social media monitor. Table \ref{isv2018:soa:indust:comptable} in Annex \hyperref[isv2018:annex:commercial]{I} offers a detailed comparative of the tools analyzed. We focus our analysis on the sources where information is gathered on, their tracking capabilities, the processing of multilingual information, and the data visualization.

\textbf{Iconoce\footnote{\url{http://info.iconoce.com/}}} is a system oriented to reputation management, offering various features such as measuring impact of campaigns, or reputation monitoring. Although it also can monitor social media (Twitter and Facebook) its strength lies on the analysis of digital press. Multilingual information can be gathered but no linguistic processing is performed (lemmatization or crosslingual searches). It has 3 separated search engines for authors, mentions and comments. A customizable dashboard offers various visualizations and data aggregations (e.g., salient term and topics, influencer, sentiment or trends). Periodical reports and alerts in the face of tendency changes are provided. As a distinctive feature, it offers a personalized press archive based on the customer configuration.

In a similar way, \textbf{INNGUMA\footnote{\url{https://www.innguma.com}}} is a tool providing business intelligence services. They put their main effort in the crawling step. Rather than offering to the user results over analysed data, the tool is designed for a group of customers to analyze the data collaboratively. Customers are provided with a search engine (more or less powerful depending on the pricing plan), and interface where they can store and share their findings.

\textbf{Lexalitycs\footnote{\url{https://www.lexalytics.com}}} and \textbf{Meaning Cloud\footnote{\url{https://www.meaningcloud.com/}}} are text analytics enterprises. Their strength is the data analysis part rather than the monitoring of many sources. Both systems are built upon robust NLP chains. Document classification, entity extraction and aspect based sentiment analysis are performed. Sentiment Analysis is approached by means of rule-based systems based on lexicons and deep linguistic analysis, offering the possibility of custom domain adaptations. Both Lexalitics and Meaning Cloud lack a result visualization interface, limiting their outputs to Excel plugins, leaving the full analysis of the data into the user's hands. 

\textbf{Websays\footnote{\url{https://websays.com/}}} monitors a wide range of sources including  news, Blogs/RSS, Forums, Facebook, Twitter, LinkedIn, Instagram, Foursquare, Pinterest, Youtube, Vimeo, Reviews (Tripadvisor, Booking,...). The user is able to configure the crawling using keywords. Negative words are also allowed in order to effectively restrict the search to the desired domain. The system is able to process data in several languages, but they report to be most effective with European languages (Spanish, English, French, Italian, and Catalan). SA is performed by combining ML algorithms and human validation, so the statistical models may learn from corrected data. The user may navigate through results using a dashboard that offers multiple filtering options. Graphs, salient terms, trending topics, influencers, sentiment and trends are provided, as well as periodical alerts and reports. The interface offers the possibility to manually edit and correct the results.

Following the same concept of Websays, \textbf{Keyhole\footnote{\url{https://keyhole.co/}}} is a monitoring and analytics tool that provides trends, insights, and analysis (including sentiment) of hashtags, keywords, or accounts on Twitter and Instagram. It reports supporting data processing in a number of languages, but no details are given on the technology. User can also track web mentions, but two separate monitoring processes must be setup.

\textbf{Lynguo\footnote{\url{http://lynguo.iic.uam.es/}}} is also in the same group of Websays and Keyhole. It claims to provide support in 24 languages, although it reports full processing chain for Spanish and English\footnote{\url{http://www.iic.uam.es/en/big-data-services/digital-environment/lynguo-en/}}. NLP is done by means of ``a range of linguistic tools to cover and combine in real time the different lexical, morphological and semantic processing layers, with machine learning and deep learning models, and software architectures''\footnote{\url{http://www.iic.uam.es/en/big-data-services/customer-intelligence-environment/natural-language-processing/}}. SA includes lexicons, customizable by the user. Monitoring is configured specifying keywords and users, allowing for negative ones as well. Lynguo is also able to geolocate comments.

\textbf{Ubermetrics\footnote{\url{https://www.ubermetrics-technologies.com/}}} is one of the few platforms that monitors multimedia sources including Youtube and Vimeo, but also TV and Radio sources. According to their reports, it processes data in 40 languages. Its visualization dashboard offers customizable graphs based on multiple search criteria. Ubermetrics main objective is analyzing virality (impact) of the mentions and author profiling.

\textbf{Snaptrends\footnote{\url{http://snaptrends.com/}}} monitors social media (Twitter, Facebook, Instagram, and Pinterest). Multilingual data is handled by means of MT (80 languages to English). It uses a proprietary NLP chain for processing English data, including sentiment analysis and relevant term extraction. The main feature for filtering large volumes of information is a geolocation-based search engine, combined with keyword based searches and other filters such as data sources. With respect to visualization, it has various data aggregations, such as influencer rankings or sentiment evolution across time by geographical area. Snaptrend makes an special effort in visualizing specific data, generating mention mosaics and timelines in real time. 

Talaia shares features with many of the aforementioned commercial solutions, yet it also possess its own characteristics. With a more robust text analysis than Iconoce and INNGUMA, and a more advanced interpretation of the data than MeaningCloud and Lexalitics, Talaia is closer to tools such as Websays and Lynguo. Having keywords organized in a taxonomy allows us to provide deeper data analysis and aggregations. Moreover, Talaia is built using open source software with a strong academical background and tested against well known benchmarks. Talaia's performance is thus, verifiable.

\section{Data Collection}
\label{isv2018:dc}

The first step of a monitoring system such as Talaia is the collection of information. The Multi Source Monitor (MSM) system\footnote{\url{http://github.com/Elhuyar/MSM}} is currently able to monitor Twitter, syndication feeds and also multimedia sources such as television or radio programs. Support for other social media such as Youtube, Facebook, etc. is under development.

MSM is a keyword based crawler, which works on a set of keywords defined by the user. Rather than a list of unconnected terms, Talaia is designed to work over a hierarchy, which allows a better organization of the data for the analysis step. In this way, keywords are defined as belonging to a specific category in the taxonomy. One handicap of crawling using a keyword-based strategy is that it is often difficult to define unambiguous terms that do not capture noisy messages. In order to minimize this situation, MSM implements a number of features:

\begin{enumerate}
	\item {\bf Regular expressions} are used to define keywords. This allows to differentiate between common words and proper names, or full words and affixes (e.g., \textit{podemos} `we can' vs. \textit{Podemos} political party). These phenomena are specially frequent in social media, where language rules are often ignored.
	
	\item {\bf Language specific keywords}. A word that is a very good keyword in a language can be a source of noise in another, e.g. \textit{mendia}, `mountain' in Spanish, is unambiguously referring to `Idoia Mendia', a Basque politician, in our context, while in Basque it is clearly ambiguous. 
	
	\item {\bf Anchor terms} usually define the general topic (e.g. election campaign) to monitor. If the user specifies that a keyword requires an anchor, then in order to accept a message containing that keyword the message must also contain at least one anchor term. Anchor terms may be keywords or not.
	
	\item {\bf Long paragraphs are split} before looking for keywords in the case of messages coming from news sites. First, it looks if any keyword appears in a candidate article. If so, it looks for keywords sentence by sentence, and those sentences are considered as the message unit.
	
\end{enumerate}

\subsection*{Language Identification (LID)}
\label{isv2018:dc:crawler:lid}

LID is indispensable in order to apply the corresponding NLP analysis. LID is integrated into the crawling process as part of the MSM system. There are two main reasons for that. First, it allows us implement the aforementioned ``language specific keyword'' feature. Second, having the language identified in the first place gives us flexibility for applying the subsequent NLP tools. At the moment language identification is implemented using the library Optimaize\footnote{https://github.com/optimaize/language-detector}, combined with source specific optimizations (social media vs. feeds). 

\section{Data Analysis}
\label{isv2018:da}

The data analysis is mainly performed by EliXa \citep{elixa-semeval2015} which integrates the following processes, each of them further detailed in the next sections.

EliXa\footnote{https://github.com/Elhuyar/Elixa} is a supervised Sentiment Analysis system. It was developed as a modular platform which allows to easily conduct experiments by replacing the modules or adding new features. It was first tested in the ABSA 2015 shared task at SemEval workshop \citep{semeval-2015-absa}. EliXa currently offers resources and models for 4 languages: Basque, Spanish, English and French. Its implementation is easily adaptable to other languages, requiring a polarity lexicon and/or a training dataset for each new language.


\subsection{Normalization}
\label{isv2018:da:pre}

To address the particular characteristics of tweets, EliXa integrates a microtext normalization module which is applied to social media messages, based on \citet{saralegi_tweetnorm_2013}. The normalizer is based on heuristic rules, such as standardizing URLs, normalizing character repetitions or dividing long words (e.g. \#AVeryLongDay $\rightarrow$ `a very long day'). Also Out Of Vocabulary (OOV) term normalization is addressed by means of language specific frequency lists based on Twitter corpora. 

Furthermore, EliXa's normalization component also includes various specific functionalities related to SA:
\begin{itemize}
	\item Emoticons are normalized into a 7 sentiment scale: \emph{smiley, crying, shock, mute, angry, kiss, sadness}.
	\item Expressions that are meaningful for detecting SA such as interjections and onomatopoeia are marked.
\end{itemize}   

Those normalized terms must be included in the polarity lexicons in order to have a greater impact in the sentiment analysis classification. Table \ref{isv2018:da:pre:rsrc-table} presents the resources provided for normalization according to their use. Word form dictionaries are composed of word forms extracted from corpora. When applying microtext normalization, candidates are compared to forms in the dict in order to discard noisy candidates. For example, if we were to normalize ``happppy'', we would know the that the correct normalization is ``happy'' by looking at theses dictionaries. 

OOV dictionaries are composed of ``OOV - standard form'' pairs. These resources are valuable to normalize slang and commonly used abbreviations. In order to produce such dictionaries word form frequency lists were generated from Twitter corpora, and after pruning standard dictionary forms, the most frequent $n$ forms were manually reviewed and manually translated\footnote{$n$ varies depending on the size of the input corpus. We reviewed up to 1,500 candidates.}. When available, dictionaries were completed using precompiled lists existing in the Web.  

Emoticon lexicon is a dictionary of regular expression matching a number of emoticons to their corresponding sentiment in the aforementioned scale.

Lastly, stopword lemma lists are used to discard most frequent lemmas when extracting n-gram features from texts. We adapted this lists to SA requirements by removing some lemmas, because of their relevance to polarity classification (e.g., no, good, ...).

\begin{table*}[ht]
	\begin{center}
		\begin{tabular}{p{0.18\textwidth}|p{0.265\textwidth}|cccc}
			\hline
			{\bf Resource} & {\bf Use} & \multicolumn{4}{c}{{\bf Language}}  \\
			&  & eu & es & en & fr\\
			\hline
			Word form dictionaries & text normalization (e.g. 4ever$\rightarrow$forever) & 122,085 & 556,501 & 67,811 & 453,037 \\ 
			OOV dictionaries & text normalization (e.g. 4ever$\rightarrow$forever) & 63 & 7,823 & 223 & 279 \\ 
			Emoticon lexicon & Polarity tagging & \multicolumn{4}{c}{60 (regexes matching emoji groups)} \\ 
			Stopword lemma lists & Polarity tagging feature extraction & 56 & 46 & 75 & 100\\
			\hline
		\end{tabular}  
		\caption{Resources for text normalization included in EliXa.}
		\label{isv2018:da:pre:rsrc-table}
	\end{center}
\end{table*}

\subsection{NLP pre-processing}
\label{isv2018:da:ixa-pipes}

EliXa currently performs tokenization, lemmatization and POS tagging prior to sentiment analysis classification. No entity recognition is applied; entities are matched only if they are defined as keywords. Although EliXa is able to work with corpora preprocessed with other taggers, its default NLP processing is made by means of IXA pipes \citep{agerri2014ixapipes} which is integrated as a library. 

\subsection{Sentiment Analysis}
\label{isv2018:da:elixa}

EliXa's core feature is its polarity classifier, which implements a multiclass Support Vector Machine (SVM) algorithm \citep{hall_weka_2009} combining the information extracted from polarity lexicons with linguistic features obtained from the NLP pre-processing step. Main features include polarity values from general and domain specific polarity lexicons, lemma and POS tag ngrams and positivity and negativity counts based on polarity lexicons. Features representing other linguistic phenomena such as treatment of negation, locutions or punctuation marks are also included. Finally, there are some social media specific features, such as the proportion of capitalized symbols (which often is used to increase the intensity of the message) or emoticon information.

EliXa currently provides ready to use polarity classification models, although one of its strengths is that new models can easily be trained if training data is available for a new domain.

\subsection{User profiling}
\label{isv2018:da:profiling}

Talaia is also capable of providing deeper analysis of the data, by means of user profiling. Specifically, geolocation, gender detection and user community identification are implemented.

Opinions gathered are geolocated. Geolocation may be done using two different approaches: (i) building a census of twitter users on a region or (ii) trying to geolocate the origin of the users detected. Approach (i) is most suitable when monitoring is done for a specific region, and high precision is required from geolocation. Details on this approach are given in section \ref{isv2018:success:pol}.

Approach (ii) allows Talaia to analyze the differences in opinions with respect to a topic that may arise between regions or countries. Geolocation is done by exploiting social media information from both messages and authors. If a message is geolocated, its information is used straightforwardly. Otherwise, user profile information is analyzed. The task is challenging, because users do not provide such information always (40\%-45\% of the users in our datasets have location information), or they define fictitious locations (e.g., `Middle earth', `In a galaxy far, far away...'; 14\%). Several geocoding APIs\footnote{OpenStreetMap Nominatim (\url{https://wiki.openstreetmap.org/wiki/Nominatim}) and Google Geocoding API \url{https://developers.google.com/maps/documentation/geocoding/intro}).} are queried, and results are then weighted, because APIs show divergent results when feeding fictitious locations. The weighted system obtains 82\% accuracy for those users containing location information in their profile. Roughly, we are able to geolocate correctly around 32\% of the users that appear in a monitoring process.

Gender detection is another important factor in many social science studies. A supervised gender classifier is implemented to infer user gender, based on features extracted from academic papers \citep{kokkos2014robust,rangel2017overview}. User gender detection is based on classifying messages, no user profile information is used. 


\section{Data Visualization}
\label{isv2018:da:vis}

The GUI has been developed using the Django Web Application framework\footnote{\url{https://www.djangoproject.com/}}. This interface provides data analysis visualizations and manages the communication with both the crawler and EliXa.

Talaia implements a number of visualizations which may be customized depending on the needs of the specific use case at hand\footnote{Existing demos and installations implement different visualizations. Visualizations for the use cases described in this paper can be seen at \url{http://talaia.elhuyar.eus/demo_eae2016}  and \url{http://behagune.elhuyar.eus/}}. The main visualizations include popularity, sympathy and antipathy comparison, evolution of mentions across time, most recent mentions, most widespread mentions, most active users in social media and news sources, and most frequent topics. All those visualizations include interactions that provide further analysis such as looking at the specific data regarding an specific party or candidate, or filtering the data according to various criteria such as language, time period, data source or author influence. All graph visualizations are implemented using d3.js\footnote{\url{https://d3js.org/}} javascript library.

The interface also has management capabilities which allows to manually review the automatic sentiment labelling. Keyword hierarchy and new website sources can be also set up through the interface. These functionalities ease the process of creating training datasets and adapting Talaia to new domains.

One of the main challenges the interface has to face is how to access data and maintain adequate time responses as the amounts of data gathered escalate. This depends to a great extent on the database optimization, but also on the number of visualizations offered by default. Talaia relies on a Mysql database. The interface does not access data from the actual tables, but rather from a joint view which is refreshed periodically. This reduces the time response from minutes to seconds \footnote{Queries retrieving a million results could take up to 7 minutes (depending on the visualizations required)), while using a joint view takes 40 seconds for the same configuration.} 

\section{Success Cases}
\label{isv2018:success}

In this section we present two real use cases where Talaia has been applied, and use them for evaluation purposes. The first one focuses on tracking cultural events. The second one analyses citizen opinions with respect to political parties and candidates during an electoral campaign. 

Both monitoring processes presented here ran on their own dedicated servers. We provide details on hardware specifications and volumes of data processed in the following subsections. As a measure of the performance capabilities of our system, the largest monitoring process we have carried out until now gathered 24M tweets per month, with an average of 700K tweets processed per day and a maximum of 1.35M tweets in a single day. Talaia ran on a server with two Intel Xeon 4 core processors (E5530) at 2.4 GHz and 16GB RAM. MySQL databases are locally stored in the server. The crawler and the text analyzers all ran locally, but no interface was implemented in this case.

\subsection{Cultural Domain}
\label{isv2018:success:cult}

Talaia was first applied in the Behagunea\footnote{\url{http://behagune.elhuyar.eus}} project. The objective of the project involved tracking the social media impact of cultural events and projects carried out (more than 500) in the framework of the Donostia European Capital of Culture (DSS2016) year during 2016. The project included monitoring opinions in press and social media in four languages: Basque, French and Spanish as coexisting languages in the different Basque speaking territories and English as international language. 

Domain adapted polarity models were created. Since events related to DSS2016 were already programmed during 2015, a previous crawl was carried out in order to build datasets. Those datasets were manually annotated for polarity in a three category scale (positive, negative, neutral). Section \ref{isv2018:eval:datasets} gives more details about the various language and domain specific datasets. Polarity classification models for the cultural domain were trained using those datasets and are distributed as part of EliXa. Section \ref{isv2018:eval:cult} gives details related to those classifiers.

Talaia ran on an Amazon AWS t4.large dedicated instance\footnote{Especications are 2 vCPUs, 8GB RAM, 100GB EBS storage disk. More information at \url{https:
	//aws.amazon.com/ec2/instance-types/}}. The crawler, the text analyzers and the interface all run locally. A total amount of 166K tweets and press mentions were gathered, with a maximum of 6.6K mentions in a single day. The interface was public and offered real-time results refreshed each 15 minutes. We can see from the volume of the data, that this was a low latency monitoring. Even if there were a lot of events to track, the local nature of most of them explains the little impact they have in social media.

\subsection{Political Domain}
\label{isv2018:success:pol}

Talaia was used to track citizen opinions during the electoral Basque electoral campaign of September 2016. The crawling was carried out during the election campaign period, starting on September 8th (23:59pm) and finishing on September 23th (23:59pm). It offers useful insights for political analysis such as sympathy rankings, the evolution of the opinions over time, most relevant messages, etc. 

The system ran 	on a server with a Intel Xeon 4 core processor (E5530) at 2.4 GHz and 16GB RAM. MySQL databases are locally stored in the server. The crawler, text analyzers and the interface all run locally. A total amount of 4.25M tweets and press mentions were gathered, with an average of 125K mentions per day, and a maximum of 433K mentions in a single day. The interface was public and offered real-time results. 

The crawler was configured to find mentions talking about the main political parties present on the campaign and their respective candidates (only main candidates monitored, i.e., those opting to be \textit{Lehendakari}, `head of the government').

Regarding social media, Twitter was monitored. Since we are talking about monitoring an event happening on a regional scope, two main restrictions were applied: only mentions written in Basque and Spanish were crawled, because those are the two official languages in the region. The second restriction was to constrain mentions to users from the specific geographical area of the Basque Country. The task was then to discard noisy messages, that do not belong to citizens involved in the election, but were likely to be talking about it. In this case, for example the crawling process was likely to capture many mentions from other regions in Spain. 

In order to solve this problem we created a census of Twitter users of the Basque Country. We developed a five step algorithm: 
\begin{enumerate}[label=(\roman*)]
	\item We gathered geolocated tweets from the Basque Country area for a certain period of time.
	\item Authors of those tweets were manually tagged with binary labels, as belonging to the required geographical area or not. Let's call this dataset $D_{geo}$.
	\item Taking users tagged as Basque citizens from the previous step, we extended our dataset by retrieving up to the first 5,000 followers and friends from each user using the twitter following API\footnote{\url{https://developer.twitter.com/en/docs/accounts-and-users/follow-search-get-users/api-reference/get-users-show}}. We compute the frequency of coocurrence for each of the candidates\footnote{The number of times a candidate appears as a follower or friend of another candidate.}, and manually label the most frequent 10,000 candidates. Let's call this dataset $D_{geo+ff-manual}$. 	
	\item We train a binary SVM classifier with a linear kernel over $D_{geo+ff-manual}$. Features of the classifier are the number of followers and friends a user has and the relative number of followers and friends (with respect to the total number of follower and friends). The classifier obtains 96\% accuracy in a 4-fold cross validation. 
	\item Repeat step (ii) with the users labelled as Basque in $D_{geo+ff-manual}$, but this time label the most frequent 20,000 candidates with the classifier trained in step (iv). Our final census consists of 23,195 user ids.
\end{enumerate}

As for the news sources, a list of 30 sources was manually compiled, including TV, printed media and radio stations, all of them with working within the regional scope.

\section{Evaluation}
\label{isv2018:eval}

For the evaluation of Talaia, we evaluate the performance of Elixa's polarity classifier for the two aforementioned domains. In all cases the L2-loss SVM implementation of the LIBLINEAR \citep{fan2008liblinear} toolkit was used as classification algorithm within Weka \citep{hall_weka_2009} data mining software. Experiments with polynomial kernels were also conducted (degrees 2-5) but we found no improvement at the expense of much longer training times. All classifiers presented in the following sections were evaluated by 10-fold cross validation. The Complexity parameter was optimized ($C=0.1$).

For the sake of comparison, all the systems presented from here onwards have been trained using the following set of features: 

\begin{itemize}
	\item 1-gram word forms with frequency $>=2$ and document frequency (df) $>=2$.
	\item POS tag 1-gram features.
	\item Polarity lemmas included in language dependent polarity lexicons. Default lexicons provided with EliXa were used (see Table \ref{isv2018:eval:tabLexicons} for details).
	\item Sentence length.
	\item Upper case ratio: percentage of the capital letters with respect to the total number of characters in a sentence.
\end{itemize} 

Microtext normalization features (URL standardization, OOV normalization and emoticon mapping) are applied before extracting the features of each sentence. 

\begin{table*}[ht]
	\begin{center}
		\begin{tabular}{p{0.12\textwidth}|p{0.38\columnwidth}|p{0.12\columnwidth}p{0.12\columnwidth}p{0.12\columnwidth}}\hline
			{\bf Language} & {\bf Lexicon} & {\bf \#neg. entries} & {\bf \#pos. entries} & {\bf Total entries} \\
			\hline
			eu & $ElhPolar_{eu}$ \citep{SANVICENTE16.468} & 742 & 499 & 1,241 \\
			es & $ElhPolar_{es}$ \citep{Saralegi2013b} & 3,314 & 1,903 & 5.217 \\
			en & $EliXa_{en}$ \citep{elixa-semeval2015} & 6,123 & 3,992 & 10,115  \\
			fr & Feel\citep{abdaoui2017feel} & 5.717 & 8,430 & 14,147 \\
			\hline
		\end{tabular}  
		\caption{Polarity lexicons used in our experiments.}
		\label{isv2018:eval:tabLexicons}
	\end{center}
\end{table*}

\subsection{Datasets}
\label{isv2018:eval:datasets}

Table \ref{isv2018:eval:datasets:table-cult} presents the statistics and class distributions of the datasets gathered and annotated in order to build the polarity classifiers for each language in the cultural domain. All annotations were done manually. Polarity was annotated at mention level. Because of the level of specificity reached when defining the keyword taxonomy, we rarely find a mention referring to more than one entity or event. Statistics show that corpora in all languages have a similar distribution, with a high number of neutral mentions, and a larger presence of positive opinions than negative ones.

\begin{table*}[ht]
	\begin{center}
		\begin{tabular}{c|cccc}
			\hline
			{\bf Language} & {\bf Total size} & \textbf{\#pos} & \textbf{\#neg} & \textbf{\#neu} \\
			\hline
			eu & 2937 & 931 & 408 & 1598 \\
			es & 4754 & 1487 & 1303 & 1964 \\
			en & 12,273 & 4,654 & 1,837 & 5,782 \\
			fr & 11,071 & 3,459 & 2,618 & 4,994 \\
			\hline
		\end{tabular}  
		\caption{Multilingual dataset statistics for the cultural domain.}
		\label{isv2018:eval:datasets:table-cult}
	\end{center}
\end{table*}

Table \ref{isv2018:eval:datasets:table-pol} shows the characteristics of the political domain datasets. In this case, each tweet was annotated with respect to a number of entities appearing in the tweet. Annotators were asked to annotate the polarity of a tweet from the perspective of each of the entities detected in a tweet, that is, a tweet may contain more than one polarity annotation. Example \ref{isv2018:eval:datasets:example} shows a real case where a tweet was given two different annotations, one for each entity (negative expressions underlined, positive ones in bold). In fact, the numbers in table \ref{isv2018:eval:datasets:table-pol} give 1.3 and 1.24 average annotations per tweet for Basque and Spanish, respectively.

\begin{adib}
	\underline{@pnvgasteiz} erabat ados, lotsagarria. Aukera ona aurrera begiratu ta \textbf{@ehbildu}|ren euskara arloko proposamena martxan jartzeko \#herriakordioa \transen{@pnvgasteiz totally agrees, shameful. Good chance to look forward and apply the proposal of @ehbildu in the field of Basque \#herriakordioa }
\label{isv2018:eval:datasets:example}	
\end{adib}

Annotating tweets in the political domain proved to be a rather challenging task. Sarcasm is often present, interpellations to a person are frequent even if they are not the target of the opinion, an opinion may be present but in an implicit manner, or a third party negative opinion may be expressed towards an entity but the author may defend it against the expressed opinion. The full annotation guidelines can be consulted in Annex \hyperref[isv2018:annex:guidelines]{II}. 

Annotation was carried out in real time during the period of the electoral campaign. We established three shifts a day to annotate messages gathered until then. Three annotators took part in the process. Because of the limited resources and the volume of messages crawled daily, each tweet was annotated by a single annotator.    

\begin{figure}[!htb]
	\minipage{0.48\textwidth}
	\includegraphics[width=\linewidth]{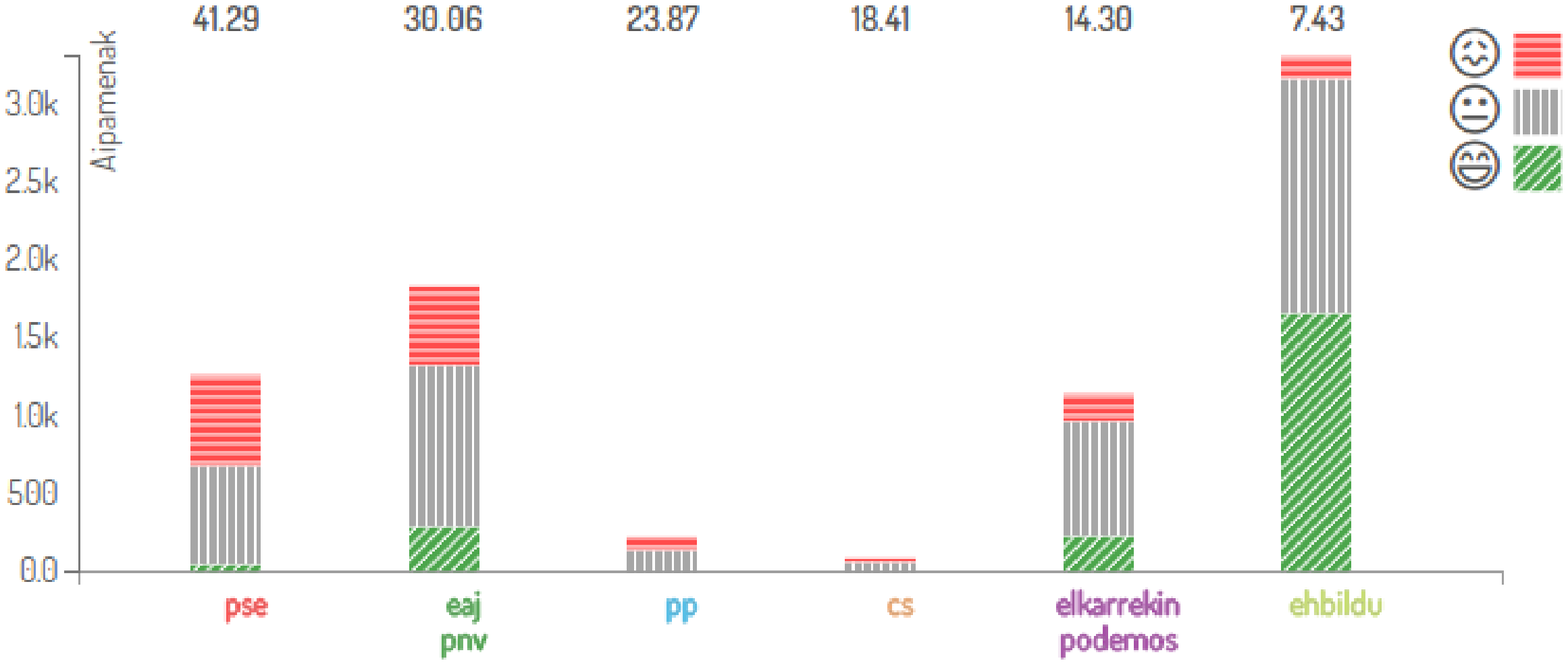}
	\caption{Distribution of mentions in Basque with respect to the political parties. From left to right, parties are sorted according the percentage of negative opinions received, with respect to the total amount of mentions received.}
	\label{talaia:fig:neg_eu}
	\endminipage\hfill
	\minipage{0.48\textwidth}
	\includegraphics[width=\linewidth]{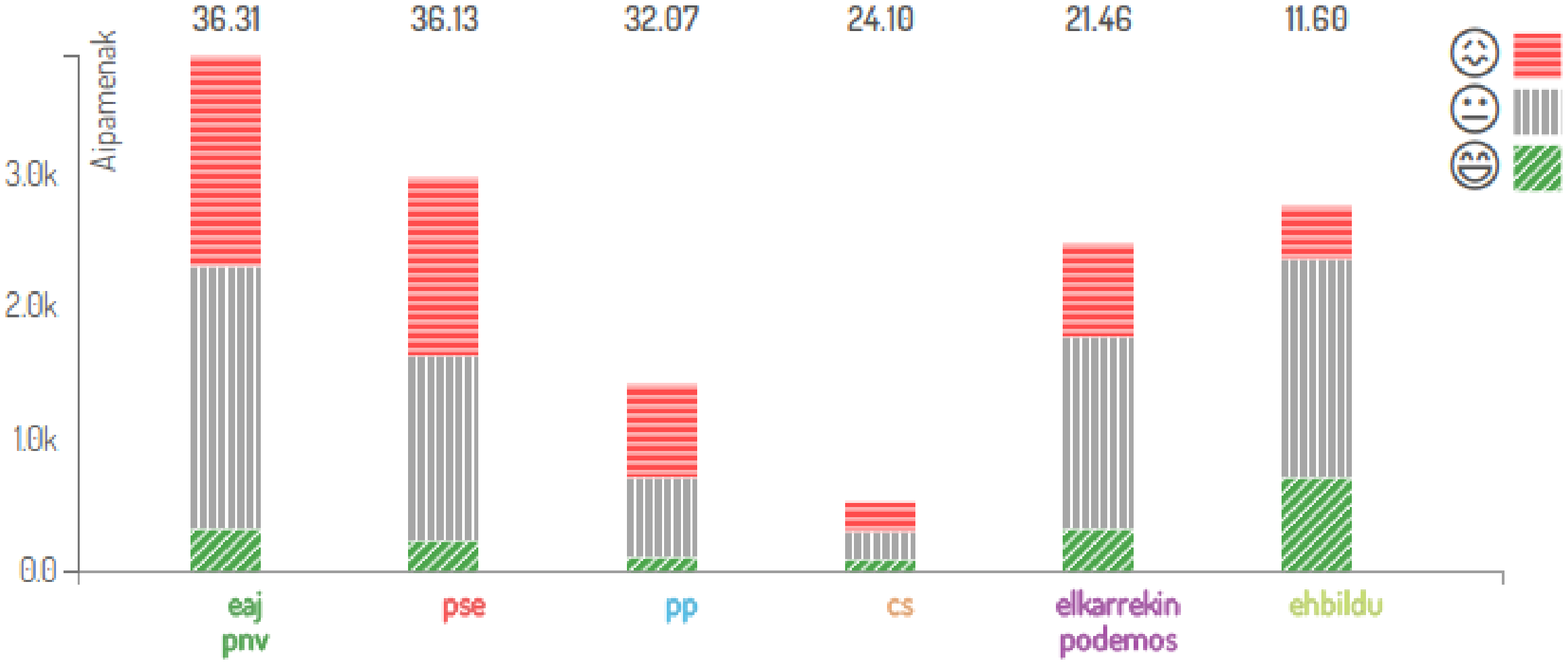}
	\caption{Distribution of mentions in Spanish with respect to the political parties. From left to right, parties are sorted according the percentage of negative opinions received, with respect to the total amount of mentions received.}
	\label{talaia:fig:neg_es}
	\endminipage\hfill
\end{figure}

Political domain datasets show very different distributions across languages. While the Basque dataset seems to follow the same pattern seen in the cultural domain, the Spanish dataset has a very high number of negative opinions. Analyzing some result samples, we realized that there are various phenomena that could explain this behaviour. First, much more debate and criticism takes place in Spanish compared to Basque where the tendency is to write a lot more supportive messages. Those not associated with Basque nationalist ideologies mainly communicate in Spanish. A clear example is that the left party EH Bildu has very few negative mentions in Basque (See figure \ref{talaia:fig:neg_eu}). Also the fact that the right wing parties such as \emph{Partido Popular} (PP) and \emph{Ciudadanos} (Cs) receive almost no attention in Basque is a symptom of the little engagement they show in this language. 

Second, left wing people were more active in Twitter (in this specific campaign), with right wing parties concentrating the largest amount of negative mentions (see figures \ref{talaia:fig:neg_es}).

Lastly, there is the effect of negative campaigning \citep{skaperdas_grofman_1995}, which is more pronounced in Spanish, because as we already said there is much more debate than in Basque. Figure \ref{talaia:fig:neg_es} also aligns with studies of negative campaigning in multi-party scenarios \citep{walter2014choosing,haselmayer2017sentiment}, being the front-runner \emph{Partido Nacionalista Vasco} (PNV) and its previous government partner \emph{Partido Socialista de Euskadi} (PSE) those who receive the greatest amount of negative mentions.

\begin{table*}[ht]
	\begin{center}
		\begin{tabular}{c|ccccc}
			\hline
			{\bf Language} & {\bf \#Tweets} & {\bf \#Annotations} & \textbf{\#pos} & \textbf{\#neg} & \textbf{\#neu} \\
			\hline
			eu & 9,418 & 11,692 & 3,974 & 3,185 & 4,533\\
			es & 15,550	& 20,278 & 3,788 & 7,601 & 8,889\\
			\hline
		\end{tabular}  
		\caption{Multilingual dataset statistics for the political domain.}
		\label{isv2018:eval:datasets:table-pol}
	\end{center}
\end{table*}

\subsection{Results}
\label{isv2018:eval:cult}

Table \ref{isv2018:eval:table} shows the performance of the various multilingual classifiers trained. 

Reported results are in general higher for the cultural domain, even if the datasets are smaller in comparison. Basque and Spanish classifier obtain results above 70\%. English and French achieve lower results. Positive mentions present the greatest challenge for English. The main reason for this is the lack of positive training examples. Neutral mentions perform very good in all languages. After analyzing a random sample, we conclude that neutral mentions are of an homogeneous nature, mainly containing agenda events or promotion messages. If we add this to the fact that neutral is the class with the highest number of examples, it seems logical that our classifiers find highly representative features for this class. 

\begin{table*}[ht]
	\begin{center}
		\begin{tabular}{cc|cccc}
			\hline
			{\bf Language} & 	\textbf{\#features} &	\textbf{acc} & \textbf{fpos} & \textbf{fneg} & \textbf{fneu} \\
			\hline
			\multicolumn{2}{c|}{\textit{Cultural Domain}} & & & &  \\
			eu & 4,777 & 74.02 & 0.658 & 0.635 & 0.803\\
			es & 10,037	& 73.03 & 0.683 & 0.756 & 0.744\\
			en & 24,183	& 70.43 & 0.715 & 0.530 & 0.743\\
			fr & 23,779	& 66.17 & 0.600 & 0.617 & 0.721\\
			\hline
			\multicolumn{2}{c|}{\textit{Political Domain}} & & & &  \\
			eu & 9,394 & 69.88 & 0.714 & 0.702 & 0.683\\
			es & 15,751 & 67.05 & 0.545 & 0.693 & 0.700\\
			\hline
			
		\end{tabular}  
		\caption{EliXa polarity classification results.}
		\label{isv2018:eval:table}
	\end{center}
\end{table*}

Regarding the political domain, if we compare Basque and Spanish classifiers, their performance drops around 4\% with respect to the results in the cultural domain. Results are not directly comparable, because political domain classifiers are evaluated over entity level tags. Also, political data is more challenging in terms of the linguistic phenomena used. We have detected a fair amount of messages containing sarcasm or opinion ambiguity towards targets.
  
In the case of Basque, the most sensible drop happens with neutral mentions. After analyzing a random sample we found that they are more heterogeneous than those in the cultural domain. They do contain agenda and promotion messages, but also many third party statements (candidate x says "...") or messages that interpellate parties and candidates over hot topics in the campaign (e.g. ``@DanielMaeztu @ehbildu @PodemosEuskadi\_ obra gelditzea onuragarria liteke ekonimia arloan 4.000 miloi gastatu eta gero?''\transen{@DanielMaeztu @ehbildu @PodemosEuskadi\_ stopping the construction would be beneficial after spending 4,000 millions?}). Many neutral messages contain personal opinions not involving any of our predefined target entities, even if they are interpellated. These phenomena make neutral class harder to represent.   

Regarding Spanish, performance for positive mentions is significantly lower. Error analysis shows that incorrectly classified instances do not fall into a single category (42\% negative, 58\% neutral). Analysing the errors, we find two main reasons. First, many positive mentions are incorrectly classified as negative because their content is mainly negative (e.g. `` @AgirreGarita La diferencia es clara, PNV apoyando el desahucio y EHBILDU al desahuciado. NO SEAS COMPLICE, no votes a quien desahucia.'' \transen{@AgirreGarita The difference is that PNV is in favour of evictions and EHBILDU is with the evicted ones. DO NOT BE AND ACCOMPLICE, don't vote to those who practice evictions.}). Our strategy for assigning message level polarity to all entities involved in a mention is prone to this type of errors. Second, as we saw for Basque, neutral mentions also contain polar expressions or opinions, making it harder to distinguish them from actual positive or negative messages.    

\section{Conclusion and Future Work}
\label{isv2018:conc}

We have presented Talaia, a real time monitor of social media and digital press. Talaia is able to extract information related to an specific topic and analyse it by means of natural language processing technologies. Two success cases and the resources generated from those cases have been described. In that sense, we have shown the ability to adapt our system to different domains and languages.

Talaia is still under development. The short term objectives include work on optimizing the information extraction process. Specifically, extracting keywords from the data downloaded up to a certain point would allow us automatically adapt the system to new terms, without losing information because the keyword hierarchy is outdated or the topic is poorly defined. 

Another important point is the adaptation of our sentiment analysis model to new domains. In that sense experiments are being carried out in order to minimize the domain adaptation process, both in terms of data collection and annotation effort.

Multilinguality is one of the main challenges of such a system. Currently the system is able to process data in 4 languages, and we are working to extend it to new languages.

Furthermore, data analysis may include further processing other than sentiment analysis. Geolocation based analysis, user community detection and other useful tasks for user profiling (e.g. gender detection) are the focus of our ongoing work. 

All the software behind the platform including the crawler, data processing chain and interface is publicly available under the GNU GPLv3 license.

\section*{Acknowledgements}

This work has been supported by the following projects: Elkarola project (Elkartek grant No. IE-14-382), and Tuner project (MINECO/FEDER grant No. TIN2015-65308-C5-1-R).

\clearpage



\bibliography{talaia_eaai}

\begin{thebibliography}{72}
\expandafter\ifx\csname natexlab\endcsname\relax\def\natexlab#1{#1}\fi
\providecommand{\url}[1]{\texttt{#1}}
\providecommand{\href}[2]{#2}
\providecommand{\path}[1]{#1}
\providecommand{\DOIprefix}{doi:}
\providecommand{\ArXivprefix}{arXiv:}
\providecommand{\URLprefix}{URL: }
\providecommand{\Pubmedprefix}{pmid:}
\providecommand{\doi}[1]{\href{http://dx.doi.org/#1}{\path{#1}}}
\providecommand{\Pubmed}[1]{\href{pmid:#1}{\path{#1}}}
\providecommand{\bibinfo}[2]{#2}
\ifx\xfnm\relax \def\xfnm[#1]{\unskip,\space#1}\fi
\bibitem[{Abdaoui et~al.(2017)Abdaoui, Az{\'e}, Bringay \&
  Poncelet}]{abdaoui2017feel}
\bibinfo{author}{Abdaoui, A.}, \bibinfo{author}{Az{\'e}, J.},
  \bibinfo{author}{Bringay, S.}, \& \bibinfo{author}{Poncelet, P.}
  (\bibinfo{year}{2017}).
\newblock \bibinfo{title}{Feel: a french expanded emotion lexicon}.
\newblock {\it \bibinfo{journal}{Language Resources and Evaluation}\/},  {\it
  \bibinfo{volume}{51}\/}, \bibinfo{pages}{833--855}.
\bibitem[{Abel et~al.(2012)Abel, Hauff, Houben, Stronkman \&
  Tao}]{abel2012twitcident}
\bibinfo{author}{Abel, F.}, \bibinfo{author}{Hauff, C.},
  \bibinfo{author}{Houben, G.-J.}, \bibinfo{author}{Stronkman, R.}, \&
  \bibinfo{author}{Tao, K.} (\bibinfo{year}{2012}).
\newblock \bibinfo{title}{Twitcident: fighting fire with information from
  social web streams}.
\newblock In {\it \bibinfo{booktitle}{Proceedings of the 21st International
  Conference on World Wide Web}\/} (pp. \bibinfo{pages}{305--308}).
\newblock \bibinfo{organization}{ACM}.
\bibitem[{Abilhoa \& De~Castro(2014)}]{abilhoa2014keyword}
\bibinfo{author}{Abilhoa, W.~D.}, \& \bibinfo{author}{De~Castro, L.~N.}
  (\bibinfo{year}{2014}).
\newblock \bibinfo{title}{A keyword extraction method from twitter messages
  represented as graphs}.
\newblock {\it \bibinfo{journal}{Applied Mathematics and Computation}\/},  {\it
  \bibinfo{volume}{240}\/}, \bibinfo{pages}{308--325}.
\bibitem[{Agerri et~al.(2014)Agerri, Bermudez \& Rigau}]{agerri2014ixapipes}
\bibinfo{author}{Agerri, R.}, \bibinfo{author}{Bermudez, J.}, \&
  \bibinfo{author}{Rigau, G.} (\bibinfo{year}{2014}).
\newblock \bibinfo{title}{Ixa pipeline: Efficient and ready to use multilingual
  nlp tools}.
\newblock In {\it \bibinfo{booktitle}{Proceedings of the 9th Language Resources
  and Evaluation Conference (LREC2014)}\/} (pp. \bibinfo{pages}{26--31}).
\newblock \bibinfo{address}{Reykjavik, Iceland}.
\bibitem[{Ahmed et~al.(2018)Ahmed, Bath, Sbaffi \&
  Demartini}]{ahmed2018measuring}
\bibinfo{author}{Ahmed, W.}, \bibinfo{author}{Bath, P.~A.},
  \bibinfo{author}{Sbaffi, L.}, \& \bibinfo{author}{Demartini, G.}
  (\bibinfo{year}{2018}).
\newblock \bibinfo{title}{Measuring the effect of public health campaigns on
  twitter: The case of world autism awareness day}.
\newblock In {\it \bibinfo{booktitle}{International Conference on
  Information}\/} (pp. \bibinfo{pages}{10--16}).
\newblock \bibinfo{organization}{Springer}.
\bibitem[{Aiello et~al.(2013)Aiello, Petkos, Martin, Corney, Papadopoulos,
  Skraba, G{\"o}ker, Kompatsiaris \& Jaimes}]{aiello2013sensing}
\bibinfo{author}{Aiello, L.~M.}, \bibinfo{author}{Petkos, G.},
  \bibinfo{author}{Martin, C.}, \bibinfo{author}{Corney, D.},
  \bibinfo{author}{Papadopoulos, S.}, \bibinfo{author}{Skraba, R.},
  \bibinfo{author}{G{\"o}ker, A.}, \bibinfo{author}{Kompatsiaris, I.}, \&
  \bibinfo{author}{Jaimes, A.} (\bibinfo{year}{2013}).
\newblock \bibinfo{title}{Sensing trending topics in twitter}.
\newblock {\it \bibinfo{journal}{IEEE Transactions on Multimedia}\/},  {\it
  \bibinfo{volume}{15}\/}, \bibinfo{pages}{1268--1282}.
\bibitem[{Alegria et~al.(2014)Alegria, Aranberri, Comas, Fresno, Gamallo,
  Padr{\'o}, San~Vicente, Turmo \& Zubiaga}]{alegria2014tweetnorm}
\bibinfo{author}{Alegria, I.}, \bibinfo{author}{Aranberri, N.},
  \bibinfo{author}{Comas, P.~R.}, \bibinfo{author}{Fresno, V.},
  \bibinfo{author}{Gamallo, P.}, \bibinfo{author}{Padr{\'o}, L.},
  \bibinfo{author}{San~Vicente, I.}, \bibinfo{author}{Turmo, J.}, \&
  \bibinfo{author}{Zubiaga, A.} (\bibinfo{year}{2014}).
\newblock \bibinfo{title}{Tweetnorm\_es corpus: an annotated corpus for spanish
  microtext normalization}.
\newblock In {\it \bibinfo{booktitle}{Proceedings of the Language Resources and
  Evaluation Conference}\/}.
\bibitem[{Alegria et~al.(2015)Alegria, Aranberri, Comas, Fresno, Gamallo,
  Padr{\'o}, San~Vicente, Turmo \& Zubiaga}]{tweetnorm_jlre_2015}
\bibinfo{author}{Alegria, I.}, \bibinfo{author}{Aranberri, N.},
  \bibinfo{author}{Comas, P.~R.}, \bibinfo{author}{Fresno, V.},
  \bibinfo{author}{Gamallo, P.}, \bibinfo{author}{Padr{\'o}, L.},
  \bibinfo{author}{San~Vicente, I.}, \bibinfo{author}{Turmo, J.}, \&
  \bibinfo{author}{Zubiaga, A.} (\bibinfo{year}{2015}).
\newblock \bibinfo{title}{Tweetnorm: a benchmark for lexical normalization of
  spanish tweets}.
\newblock {\it \bibinfo{journal}{Language Resources and Evaluation}\/},  {\it
  \bibinfo{volume}{49}\/}, \bibinfo{pages}{883--905}.
\bibitem[{Allan et~al.(1998)Allan, Carbonell, Doddington, Yamron, Yang
  et~al.}]{allan1998topic}
\bibinfo{author}{Allan, J.}, \bibinfo{author}{Carbonell, J.},
  \bibinfo{author}{Doddington, G.}, \bibinfo{author}{Yamron, J.},
  \bibinfo{author}{Yang, Y.} et~al. (\bibinfo{year}{1998}).
\newblock \bibinfo{title}{Topic detection and tracking pilot study: Final
  report}.
\newblock In {\it \bibinfo{booktitle}{Proceedings of the DARPA broadcast news
  transcription and understanding workshop}\/} (pp. \bibinfo{pages}{194--218}).
\newblock \bibinfo{organization}{Citeseer} volume \bibinfo{volume}{1998}.
\bibitem[{Barbosa \& Feng(2010)}]{barbosa_robust_2010}
\bibinfo{author}{Barbosa, L.}, \& \bibinfo{author}{Feng, J.}
  (\bibinfo{year}{2010}).
\newblock \bibinfo{title}{Robust sentiment detection on twitter from biased and
  noisy data}.
\newblock In {\it \bibinfo{booktitle}{Proceedings of the 23rd International
  Conference on Computational Linguistics: Posters}\/} {COLING} '10 (pp.
  \bibinfo{pages}{36--44}).
\newblock \bibinfo{address}{Stroudsburg, {PA}, {USA}}.
\bibitem[{Belainine et~al.(2016)Belainine, Fonseca \&
  Sadat}]{Belainine_W16-3915}
\bibinfo{author}{Belainine, B.}, \bibinfo{author}{Fonseca, A.}, \&
  \bibinfo{author}{Sadat, F.} (\bibinfo{year}{2016}).
\newblock \bibinfo{title}{Named entity recognition and hashtag decomposition to
  improve the classification of tweets}.
\newblock In {\it \bibinfo{booktitle}{Proceedings of the 2nd Workshop on Noisy
  User-generated Text (WNUT)}\/} (pp. \bibinfo{pages}{102--111}).
\bibitem[{Blei et~al.(2003)Blei, Ng \& Jordan}]{blei2003latent}
\bibinfo{author}{Blei, D.~M.}, \bibinfo{author}{Ng, A.~Y.}, \&
  \bibinfo{author}{Jordan, M.~I.} (\bibinfo{year}{2003}).
\newblock \bibinfo{title}{Latent dirichlet allocation}.
\newblock {\it \bibinfo{journal}{Journal of machine Learning research}\/},
  {\it \bibinfo{volume}{3}\/}, \bibinfo{pages}{993--1022}.
\bibitem[{Bollen et~al.(2010)Bollen, Mao \& Zeng}]{bollen_twitter_2010}
\bibinfo{author}{Bollen, J.}, \bibinfo{author}{Mao, H.}, \&
  \bibinfo{author}{Zeng, X.-J.} (\bibinfo{year}{2010}).
\newblock \bibinfo{title}{Twitter mood predicts the stock market}.
\newblock {\it \bibinfo{journal}{1010.3003}\/}, .
\bibitem[{Brody \& Diakopoulos(2011)}]{brody_cool2011}
\bibinfo{author}{Brody, S.}, \& \bibinfo{author}{Diakopoulos, N.}
  (\bibinfo{year}{2011}).
\newblock \bibinfo{title}{Cooooooooooooooollllllllllllll!!!!!!!!!!!: using word
  lengthening to detect sentiment in microblogs}.
\newblock In {\it \bibinfo{booktitle}{Proceedings of the Conference on
  Empirical Methods in Natural Language Processing}\/} {EMNLP} '11 (pp.
  \bibinfo{pages}{562--570}).
\bibitem[{Brun \& Roux(2014)}]{BrunRoux_F14-2015}
\bibinfo{author}{Brun, C.}, \& \bibinfo{author}{Roux, C.}
  (\bibinfo{year}{2014}).
\newblock \bibinfo{title}{Decomposing hashtags to improve tweet polarity
  classification (d{\'e}composition des {\guillemotleft}hash
  tags{\guillemotright} pour l'am{\'e}lioration de la classification en
  polarit{\'e} des {\guillemotleft}tweets{\guillemotright})[in french]}.
\newblock In {\it \bibinfo{booktitle}{Proceedings of TALN 2014 (Volume 2: Short
  Papers)}\/} (pp. \bibinfo{pages}{473--478}).
\bibitem[{Ceron et~al.(2015)Ceron, Curini \& Iacus}]{ceron2015using}
\bibinfo{author}{Ceron, A.}, \bibinfo{author}{Curini, L.}, \&
  \bibinfo{author}{Iacus, S.~M.} (\bibinfo{year}{2015}).
\newblock \bibinfo{title}{Using sentiment analysis to monitor electoral
  campaigns: Method matters—evidence from the united states and italy}.
\newblock {\it \bibinfo{journal}{Social Science Computer Review}\/},  {\it
  \bibinfo{volume}{33}\/}, \bibinfo{pages}{3--20}.
\bibitem[{Chen et~al.(2012)Chen, Chiang \& Storey}]{chen2012business}
\bibinfo{author}{Chen, H.}, \bibinfo{author}{Chiang, R.~H.}, \&
  \bibinfo{author}{Storey, V.~C.} (\bibinfo{year}{2012}).
\newblock \bibinfo{title}{Business intelligence and analytics: from big data to
  big impact}.
\newblock {\it \bibinfo{journal}{MIS quarterly}\/},  (pp.
  \bibinfo{pages}{1165--1188}).
\bibitem[{Cliche(2017)}]{Cliche2017}
\bibinfo{author}{Cliche, M.} (\bibinfo{year}{2017}).
\newblock \bibinfo{title}{Bb{\_}twtr at semeval-2017 task 4: Twitter sentiment
  analysis with cnns and lstms}.
\newblock In {\it \bibinfo{booktitle}{Proceedings of the 11th International
  Workshop on Semantic Evaluation (SemEval-2017)}\/} (pp.
  \bibinfo{pages}{573--580}).
\bibitem[{Dadvar et~al.(2013)Dadvar, Trieschnigg, Ordelman \&
  de~Jong}]{dadvar2013improving}
\bibinfo{author}{Dadvar, M.}, \bibinfo{author}{Trieschnigg, D.},
  \bibinfo{author}{Ordelman, R.}, \& \bibinfo{author}{de~Jong, F.}
  (\bibinfo{year}{2013}).
\newblock \bibinfo{title}{Improving cyberbullying detection with user context}.
\newblock In {\it \bibinfo{booktitle}{European Conference on Information
  Retrieval}\/} (pp. \bibinfo{pages}{693--696}).
\newblock \bibinfo{organization}{Springer}.
\bibitem[{Dai \& Le(2015)}]{DaiLe2015NIPS}
\bibinfo{author}{Dai, A.~M.}, \& \bibinfo{author}{Le, Q.~V.}
  (\bibinfo{year}{2015}).
\newblock \bibinfo{title}{Semi-supervised sequence learning}.
\newblock In \bibinfo{editor}{C.~Cortes}, \bibinfo{editor}{N.~D. Lawrence},
  \bibinfo{editor}{D.~D. Lee}, \bibinfo{editor}{M.~Sugiyama}, \&
  \bibinfo{editor}{R.~Garnett} (Eds.), {\it \bibinfo{booktitle}{Advances in
  Neural Information Processing Systems 28}\/} (pp.
  \bibinfo{pages}{3079--3087}).
\bibitem[{Deriu et~al.(2017)Deriu, Lucchi, De~Luca, Severyn, M\"{u}ller,
  Cieliebak, Hofmann \& Jaggi}]{Deriu2017}
\bibinfo{author}{Deriu, J.}, \bibinfo{author}{Lucchi, A.},
  \bibinfo{author}{De~Luca, V.}, \bibinfo{author}{Severyn, A.},
  \bibinfo{author}{M\"{u}ller, S.}, \bibinfo{author}{Cieliebak, M.},
  \bibinfo{author}{Hofmann, T.}, \& \bibinfo{author}{Jaggi, M.}
  (\bibinfo{year}{2017}).
\newblock \bibinfo{title}{Leveraging large amounts of weakly supervised data
  for multi-language sentiment classification}.
\newblock In {\it \bibinfo{booktitle}{Proceedings of the 26th International
  Conference on World Wide Web}\/} WWW '17 (pp. \bibinfo{pages}{1045--1052}).
\newblock \bibinfo{address}{Republic and Canton of Geneva, Switzerland}.
\bibitem[{Fan et~al.(2008)Fan, Chang, Hsieh, Wang \& Lin}]{fan2008liblinear}
\bibinfo{author}{Fan, R.-E.}, \bibinfo{author}{Chang, K.-W.},
  \bibinfo{author}{Hsieh, C.-J.}, \bibinfo{author}{Wang, X.-R.}, \&
  \bibinfo{author}{Lin, C.-J.} (\bibinfo{year}{2008}).
\newblock \bibinfo{title}{Liblinear: A library for large linear
  classification}.
\newblock {\it \bibinfo{journal}{Journal of machine learning research}\/},
  {\it \bibinfo{volume}{9}\/}, \bibinfo{pages}{1871--1874}.
\bibitem[{Foster et~al.(2011)Foster, Cetinoglu, Wagner, Le~Roux, Hogan, Nivre,
  Hogan \& van Genabith}]{foster_hardtoparse2011}
\bibinfo{author}{Foster, J.}, \bibinfo{author}{Cetinoglu, O.},
  \bibinfo{author}{Wagner, J.}, \bibinfo{author}{Le~Roux, J.},
  \bibinfo{author}{Hogan, S.}, \bibinfo{author}{Nivre, J.},
  \bibinfo{author}{Hogan, D.}, \& \bibinfo{author}{van Genabith, J.}
  (\bibinfo{year}{2011}).
\newblock \bibinfo{title}{\#hardtoparse: {POS} tagging and parsing the
  twitterverse}.
\newblock In {\it \bibinfo{booktitle}{Workshops at the Twenty-Fifth {AAAI}
  Conference on Artificial Intelligence}\/}.
\bibitem[{Go et~al.(2009)Go, Bhayani \& Huang}]{go_twitter_2009}
\bibinfo{author}{Go, A.}, \bibinfo{author}{Bhayani, R.}, \&
  \bibinfo{author}{Huang, L.} (\bibinfo{year}{2009}).
\newblock \bibinfo{title}{Twitter sentiment classification using distant
  supervision}.
\newblock {\it \bibinfo{journal}{{CS224N} Project Report, Stanford}\/},  (pp.
  \bibinfo{pages}{1--12}).
\bibitem[{Hall et~al.(2009)Hall, Frank, Holmes, Pfahringer, Reutemann \&
  Witten}]{hall_weka_2009}
\bibinfo{author}{Hall, M.}, \bibinfo{author}{Frank, E.},
  \bibinfo{author}{Holmes, G.}, \bibinfo{author}{Pfahringer, B.},
  \bibinfo{author}{Reutemann, P.}, \& \bibinfo{author}{Witten, I.~H.}
  (\bibinfo{year}{2009}).
\newblock \bibinfo{title}{The {WEKA} data mining software: an update}.
\newblock {\it \bibinfo{journal}{{SIGKDD} Explor. Newsl.}\/},  {\it
  \bibinfo{volume}{11}\/}, \bibinfo{pages}{10--18}.
\bibitem[{Han \& Baldwin(2011)}]{han_lexical_2011}
\bibinfo{author}{Han, B.}, \& \bibinfo{author}{Baldwin, T.}
  (\bibinfo{year}{2011}).
\newblock \bibinfo{title}{Lexical normalisation of short text messages: Makn
  sens a\# twitter}.
\newblock In {\it \bibinfo{booktitle}{Proceedings of the 49th Annual Meeting of
  the Association for Computational Linguistics: Human Language
  Technologies}\/} (pp. \bibinfo{pages}{368--378}).
\newblock volume~\bibinfo{volume}{1}.
\bibitem[{Haselmayer \& Jenny(2017)}]{haselmayer2017sentiment}
\bibinfo{author}{Haselmayer, M.}, \& \bibinfo{author}{Jenny, M.}
  (\bibinfo{year}{2017}).
\newblock \bibinfo{title}{Sentiment analysis of political communication:
  combining a dictionary approach with crowdcoding}.
\newblock {\it \bibinfo{journal}{Quality \& quantity}\/},  {\it
  \bibinfo{volume}{51}\/}, \bibinfo{pages}{2623--2646}.
\bibitem[{He et~al.(2013)He, Zha \& Li}]{he2013social}
\bibinfo{author}{He, W.}, \bibinfo{author}{Zha, S.}, \& \bibinfo{author}{Li,
  L.} (\bibinfo{year}{2013}).
\newblock \bibinfo{title}{Social media competitive analysis and text mining: A
  case study in the pizza industry}.
\newblock {\it \bibinfo{journal}{International Journal of Information
  Management}\/},  {\it \bibinfo{volume}{33}\/}, \bibinfo{pages}{464--472}.
\bibitem[{Howard \& Ruder(2018)}]{ulmfit2018}
\bibinfo{author}{Howard, J.}, \& \bibinfo{author}{Ruder, S.}
  (\bibinfo{year}{2018}).
\newblock \bibinfo{title}{Universal language model fine-tuning for text
  classification}.
\newblock In {\it \bibinfo{booktitle}{Proceedings of the 56th Annual Meeting of
  the Association for Computational Linguistics (Volume 1: Long Papers)}\/}
  (pp. \bibinfo{pages}{328--339}).
\bibitem[{Hu \& Liu(2004)}]{hu_mining_2004}
\bibinfo{author}{Hu, M.}, \& \bibinfo{author}{Liu, B.} (\bibinfo{year}{2004}).
\newblock \bibinfo{title}{Mining and summarizing customer reviews}.
\newblock In {\it \bibinfo{booktitle}{Proceedings of the tenth {ACM} {SIGKDD}
  international conference on Knowledge discovery and data mining}\/} (pp.
  \bibinfo{pages}{168--177}).
\bibitem[{Johnson \& Zhang(2016)}]{johnson2016supervised}
\bibinfo{author}{Johnson, R.}, \& \bibinfo{author}{Zhang, T.}
  (\bibinfo{year}{2016}).
\newblock \bibinfo{title}{Supervised and semi-supervised text categorization
  using lstm for region embeddings}.
\newblock In {\it \bibinfo{booktitle}{International Conference on Machine
  Learning}\/} (pp. \bibinfo{pages}{526--534}).
\bibitem[{Jurgens et~al.(2015)Jurgens, Finethy, McCorriston, Xu \&
  Ruths}]{jurgens2015geolocation}
\bibinfo{author}{Jurgens, D.}, \bibinfo{author}{Finethy, T.},
  \bibinfo{author}{McCorriston, J.}, \bibinfo{author}{Xu, Y.~T.}, \&
  \bibinfo{author}{Ruths, D.} (\bibinfo{year}{2015}).
\newblock \bibinfo{title}{Geolocation prediction in twitter using social
  networks: A critical analysis and review of current practice.}
\newblock {\it \bibinfo{journal}{ICWSM}\/},  {\it \bibinfo{volume}{15}\/},
  \bibinfo{pages}{188--197}.
\bibitem[{Kim et~al.(2016)Kim, Banchs \& Li}]{kim2016exploring}
\bibinfo{author}{Kim, S.}, \bibinfo{author}{Banchs, R.}, \&
  \bibinfo{author}{Li, H.} (\bibinfo{year}{2016}).
\newblock \bibinfo{title}{Exploring convolutional and recurrent neural networks
  in sequential labelling for dialogue topic tracking}.
\newblock In {\it \bibinfo{booktitle}{Proceedings of the 54th Annual Meeting of
  the Association for Computational Linguistics (Volume 1: Long Papers)}\/}
  (pp. \bibinfo{pages}{963--973}).
\newblock volume~\bibinfo{volume}{1}.
\bibitem[{Kokkos \& Tzouramanis(2014)}]{kokkos2014robust}
\bibinfo{author}{Kokkos, A.}, \& \bibinfo{author}{Tzouramanis, T.}
  (\bibinfo{year}{2014}).
\newblock \bibinfo{title}{A robust gender inference model for online social
  networks and its application to linkedin and twitter}.
\newblock {\it \bibinfo{journal}{First Monday}\/},  {\it
  \bibinfo{volume}{19}\/}.
\bibitem[{Kouloumpis et~al.(2011)Kouloumpis, Wilson \&
  Moore}]{kouloumpis_twitter_2011}
\bibinfo{author}{Kouloumpis, E.}, \bibinfo{author}{Wilson, T.}, \&
  \bibinfo{author}{Moore, J.~D.} (\bibinfo{year}{2011}).
\newblock \bibinfo{title}{Twitter sentiment analysis: The good the bad and the
  {OMG!}}
\newblock In {\it \bibinfo{booktitle}{Fifth International {AAAI} Conference on
  Weblogs and Social Media}\/}.
\bibitem[{Liu et~al.(2012)Liu, Weng \& Jiang}]{liu_broad-coverage_2012}
\bibinfo{author}{Liu, F.}, \bibinfo{author}{Weng, F.}, \&
  \bibinfo{author}{Jiang, X.} (\bibinfo{year}{2012}).
\newblock \bibinfo{title}{A broad-coverage normalization system for social
  media language}.
\newblock In {\it \bibinfo{booktitle}{Proceedings of the 50th Annual Meeting of
  the Association for Computational Linguistics {(Volume} 1: Long Papers)}\/}
  (pp. \bibinfo{pages}{1035--1044}).
\newblock \bibinfo{address}{Jeju Island, Korea}.
\bibitem[{Liu et~al.(2011)Liu, Zhang, Wei \& Zhou}]{liu_recognizing_2011}
\bibinfo{author}{Liu, X.}, \bibinfo{author}{Zhang, S.}, \bibinfo{author}{Wei,
  F.}, \& \bibinfo{author}{Zhou, M.} (\bibinfo{year}{2011}).
\newblock \bibinfo{title}{Recognizing named entities in tweets.}
\newblock In {\it \bibinfo{booktitle}{Proceedings of the 49th Annual Meeting of
  the Association for Computational Linguistics (ACL)}\/} (pp.
  \bibinfo{pages}{359--367}).
\bibitem[{Miao et~al.(2017)Miao, Chen, Fang, He, Zhou, Zhang \&
  Zha}]{miao2017cost}
\bibinfo{author}{Miao, Z.}, \bibinfo{author}{Chen, K.}, \bibinfo{author}{Fang,
  Y.}, \bibinfo{author}{He, J.}, \bibinfo{author}{Zhou, Y.},
  \bibinfo{author}{Zhang, W.}, \& \bibinfo{author}{Zha, H.}
  (\bibinfo{year}{2017}).
\newblock \bibinfo{title}{Cost-effective online trending topic detection and
  popularity prediction in microblogging}.
\newblock {\it \bibinfo{journal}{ACM Transactions on Information Systems
  (TOIS)}\/},  {\it \bibinfo{volume}{35}\/}, \bibinfo{pages}{18}.
\bibitem[{Mikolov et~al.(2013)Mikolov, Sutskever, Chen, Corrado \&
  Dean}]{mikolov2013distributed}
\bibinfo{author}{Mikolov, T.}, \bibinfo{author}{Sutskever, I.},
  \bibinfo{author}{Chen, K.}, \bibinfo{author}{Corrado, G.~S.}, \&
  \bibinfo{author}{Dean, J.} (\bibinfo{year}{2013}).
\newblock \bibinfo{title}{Distributed representations of words and phrases and
  their compositionality}.
\newblock In {\it \bibinfo{booktitle}{Advances in Neural Information Processing
  Systems}\/} (pp. \bibinfo{pages}{3111--3119}).
\bibitem[{Mohammad et~al.(2013)Mohammad, Kiritchenko \&
  Zhu}]{mohammad_semeval_2013}
\bibinfo{author}{Mohammad, S.}, \bibinfo{author}{Kiritchenko, S.}, \&
  \bibinfo{author}{Zhu, X.} (\bibinfo{year}{2013}).
\newblock \bibinfo{title}{{NRC}-canada: Building the state-of-the-art in
  sentiment analysis of tweets}.
\newblock In {\it \bibinfo{booktitle}{Second Joint Conference on Lexical and
  Computational Semantics (*{SEM}), Volume 2: Proceedings of the Seventh
  International Workshop on Semantic Evaluation ({SemEval} 2013)}\/} (pp.
  \bibinfo{pages}{321--327}).
\bibitem[{Mostafa(2013)}]{mostafa2013more}
\bibinfo{author}{Mostafa, M.~M.} (\bibinfo{year}{2013}).
\newblock \bibinfo{title}{More than words: Social networks’ text mining for
  consumer brand sentiments}.
\newblock {\it \bibinfo{journal}{Expert Systems with Applications}\/},  {\it
  \bibinfo{volume}{40}\/}, \bibinfo{pages}{4241--4251}.
\bibitem[{Nakov et~al.(2016)Nakov, Ritter, Rosenthal, Sebastiani \&
  Stoyanov}]{Semeval2016-Twitter-task4}
\bibinfo{author}{Nakov, P.}, \bibinfo{author}{Ritter, A.},
  \bibinfo{author}{Rosenthal, S.}, \bibinfo{author}{Sebastiani, F.}, \&
  \bibinfo{author}{Stoyanov, V.} (\bibinfo{year}{2016}).
\newblock \bibinfo{title}{Semeval-2016 task 4: Sentiment analysis in twitter}.
\newblock In {\it \bibinfo{booktitle}{Proceedings of the 10th International
  Workshop on Semantic Evaluation (SemEval-2016)}\/} (pp.
  \bibinfo{pages}{1--18}).
\bibitem[{Nguyen et~al.(2016)Nguyen, Shin \& Yoo}]{nguyen2016hot}
\bibinfo{author}{Nguyen, K.-L.}, \bibinfo{author}{Shin, B.-J.}, \&
  \bibinfo{author}{Yoo, S.~J.} (\bibinfo{year}{2016}).
\newblock \bibinfo{title}{Hot topic detection and technology trend tracking for
  patents utilizing term frequency and proportional document frequency and
  semantic information}.
\newblock In {\it \bibinfo{booktitle}{Big Data and Smart Computing (BigComp),
  2016 International Conference on}\/} (pp. \bibinfo{pages}{223--230}).
\newblock \bibinfo{organization}{IEEE}.
\bibitem[{{O'Connor} et~al.(2010){O'Connor}, Balasubramanyan, Routledge \&
  Smith}]{oconnor_tweets_2010}
\bibinfo{author}{{O'Connor}, B.}, \bibinfo{author}{Balasubramanyan, R.},
  \bibinfo{author}{Routledge, B.~R.}, \& \bibinfo{author}{Smith, N.~A.}
  (\bibinfo{year}{2010}).
\newblock \bibinfo{title}{From tweets to polls: Linking text sentiment to
  public opinion time series}.
\newblock In {\it \bibinfo{booktitle}{Fourth International {AAAI} Conference on
  Weblogs and Social Media}\/}.
\bibitem[{Oliveira et~al.(2017)Oliveira, Cortez \& Areal}]{oliveira2017impact}
\bibinfo{author}{Oliveira, N.}, \bibinfo{author}{Cortez, P.}, \&
  \bibinfo{author}{Areal, N.} (\bibinfo{year}{2017}).
\newblock \bibinfo{title}{The impact of microblogging data for stock market
  prediction: using twitter to predict returns, volatility, trading volume and
  survey sentiment indices}.
\newblock {\it \bibinfo{journal}{Expert Systems with Applications}\/},  {\it
  \bibinfo{volume}{73}\/}, \bibinfo{pages}{125--144}.
\bibitem[{Osborne et~al.(2014)Osborne, Moran, McCreadie, Von~Lunen, Sykora,
  Cano, Ireson, Macdonald, Ounis, He et~al.}]{osborne2014real}
\bibinfo{author}{Osborne, M.}, \bibinfo{author}{Moran, S.},
  \bibinfo{author}{McCreadie, R.}, \bibinfo{author}{Von~Lunen, A.},
  \bibinfo{author}{Sykora, M.}, \bibinfo{author}{Cano, E.},
  \bibinfo{author}{Ireson, N.}, \bibinfo{author}{Macdonald, C.},
  \bibinfo{author}{Ounis, I.}, \bibinfo{author}{He, Y.} et~al.
  (\bibinfo{year}{2014}).
\newblock \bibinfo{title}{Real-time detection, tracking, and monitoring of
  automatically discovered events in social media}.
\newblock {\it \bibinfo{journal}{ACL 2014}\/},  (p.~\bibinfo{pages}{37}).
\bibitem[{{\"O}zt{\"u}rk \& Ayvaz(2018)}]{ozturk2018sentiment}
\bibinfo{author}{{\"O}zt{\"u}rk, N.}, \& \bibinfo{author}{Ayvaz, S.}
  (\bibinfo{year}{2018}).
\newblock \bibinfo{title}{Sentiment analysis on twitter: A text mining approach
  to the syrian refugee crisis}.
\newblock {\it \bibinfo{journal}{Telematics and Informatics}\/},  {\it
  \bibinfo{volume}{35}\/}, \bibinfo{pages}{136--147}.
\bibitem[{Pontiki et~al.(2015)Pontiki, Galanis, Papageorgiou, Manandhar \&
  Androutsopoulos}]{semeval-2015-absa}
\bibinfo{author}{Pontiki, M.}, \bibinfo{author}{Galanis, D.},
  \bibinfo{author}{Papageorgiou, H.}, \bibinfo{author}{Manandhar, S.}, \&
  \bibinfo{author}{Androutsopoulos, I.} (\bibinfo{year}{2015}).
\newblock \bibinfo{title}{Semeval-2015 task 12: Aspect based sentiment
  analysis}.
\newblock In {\it \bibinfo{booktitle}{Proceedings of the 9th International
  Workshop on Semantic Evaluation (SemEval 2015)}\/} (pp.
  \bibinfo{pages}{486--495}).
\bibitem[{Pontiki et~al.(2014)Pontiki, Galanis, Pavlopoulos, Papageorgiou,
  Androutsopoulos \& Manandhar}]{semeval-2014_4}
\bibinfo{author}{Pontiki, M.}, \bibinfo{author}{Galanis, D.},
  \bibinfo{author}{Pavlopoulos, J.}, \bibinfo{author}{Papageorgiou, H.},
  \bibinfo{author}{Androutsopoulos, I.}, \& \bibinfo{author}{Manandhar, S.}
  (\bibinfo{year}{2014}).
\newblock \bibinfo{title}{Semeval-2014 task 4: Aspect based sentiment
  analysis}.
\newblock In {\it \bibinfo{booktitle}{Proceedings of the International Workshop
  on Semantic Evaluation ({SemEval})}\/}.
\bibitem[{Pope \& Griffith(2016)}]{griffith2016analysis}
\bibinfo{author}{Pope, D.}, \& \bibinfo{author}{Griffith, J.}
  (\bibinfo{year}{2016}).
\newblock \bibinfo{title}{An analysis of online twitter sentiment surrounding
  the european refugee crisis}.
\newblock In {\it \bibinfo{booktitle}{Proceedings of the International Joint
  Conference on Knowledge Discovery, Knowledge Engineering and Knowledge
  Management}\/} IC3K 2016 (pp. \bibinfo{pages}{299--306}).
\newblock \bibinfo{address}{Portugal}.
\bibitem[{Preo{\c{t}}iuc-Pietro \& Cohn(2013)}]{preoctiuc2013temporal}
\bibinfo{author}{Preo{\c{t}}iuc-Pietro, D.}, \& \bibinfo{author}{Cohn, T.}
  (\bibinfo{year}{2013}).
\newblock \bibinfo{title}{A temporal model of text periodicities using gaussian
  processes}.
\newblock In {\it \bibinfo{booktitle}{Proceedings of the 2013 Conference on
  Empirical Methods in Natural Language Processing}\/} (pp.
  \bibinfo{pages}{977--988}).
\bibitem[{Rangel et~al.(2017)Rangel, Rosso, Potthast \&
  Stein}]{rangel2017overview}
\bibinfo{author}{Rangel, F.}, \bibinfo{author}{Rosso, P.},
  \bibinfo{author}{Potthast, M.}, \& \bibinfo{author}{Stein, B.}
  (\bibinfo{year}{2017}).
\newblock \bibinfo{title}{Overview of the 5th author profiling task at pan
  2017: Gender and language variety identification in twitter}.
\newblock {\it \bibinfo{journal}{Working Notes Papers of the CLEF}\/}, .
\bibitem[{Rom{\'a}n et~al.(2015)Rom{\'a}n, C{\'a}mara, Morera \&
  Zafra}]{roman_tass2014}
\bibinfo{author}{Rom{\'a}n, J.~V.}, \bibinfo{author}{C{\'a}mara, E.~M.},
  \bibinfo{author}{Morera, J.~G.}, \& \bibinfo{author}{Zafra, S. M.~J.}
  (\bibinfo{year}{2015}).
\newblock \bibinfo{title}{Tass 2014-the challenge of aspect-based sentiment
  analysis}.
\newblock {\it \bibinfo{journal}{Procesamiento del Lenguaje Natural}\/},  {\it
  \bibinfo{volume}{54}\/}, \bibinfo{pages}{61--68}.
\bibitem[{Rosenthal et~al.(2017)Rosenthal, Farra \&
  Nakov}]{Semeval2017-twitter-task4}
\bibinfo{author}{Rosenthal, S.}, \bibinfo{author}{Farra, N.}, \&
  \bibinfo{author}{Nakov, P.} (\bibinfo{year}{2017}).
\newblock \bibinfo{title}{Semeval-2017 task 4: Sentiment analysis in twitter}.
\newblock In {\it \bibinfo{booktitle}{Proceedings of the 11th International
  Workshop on Semantic Evaluation (SemEval-2017)}\/} (pp.
  \bibinfo{pages}{502--518}).
\bibitem[{Rosenthal et~al.(2014)Rosenthal, Nakov, Ritter \&
  Stoyanov}]{semeval-2014_9}
\bibinfo{author}{Rosenthal, S.}, \bibinfo{author}{Nakov, P.},
  \bibinfo{author}{Ritter, A.}, \& \bibinfo{author}{Stoyanov, V.}
  (\bibinfo{year}{2014}).
\newblock \bibinfo{title}{Semeval-2014 task 9: Sentiment analysis in twitter}.
\newblock In {\it \bibinfo{booktitle}{Proceedings of the 8th International
  Workshop on Semantic Evaluation, {SemEval}}\/}.
\newblock volume~\bibinfo{volume}{14}.
\bibitem[{{San Vicente} \& Saralegi(2016)}]{SANVICENTE16.468}
\bibinfo{author}{{San Vicente}, I.}, \& \bibinfo{author}{Saralegi, X.}
  (\bibinfo{year}{2016}).
\newblock \bibinfo{title}{Polarity lexicon building: to what extent is the
  manual effort worth?}
\newblock In {\it \bibinfo{booktitle}{Proceedings of the Tenth International
  Conference on Language Resources and Evaluation (LREC 2016)}\/}.
\bibitem[{San~Vicente et~al.(2015)San~Vicente, Saralegi \&
  Agerri}]{elixa-semeval2015}
\bibinfo{author}{San~Vicente, I.}, \bibinfo{author}{Saralegi, X.}, \&
  \bibinfo{author}{Agerri, R.} (\bibinfo{year}{2015}).
\newblock \bibinfo{title}{Elixa: A modular and flexible absa platform}.
\newblock In {\it \bibinfo{booktitle}{Proceedings of the 9th International
  Workshop on Semantic Evaluation (SemEval 2015)}\/} (pp.
  \bibinfo{pages}{748--752}).
\bibitem[{Saralegi \& {San Vicente}(2012)}]{saralegi_tass_2012}
\bibinfo{author}{Saralegi, X.}, \& \bibinfo{author}{{San Vicente}, I.}
  (\bibinfo{year}{2012}).
\newblock \bibinfo{title}{Tass: Detecting sentiments in spanish tweets}.
\newblock In {\it \bibinfo{booktitle}{Proceedings of the TASS Workshop at
  SEPLN}\/}.
\bibitem[{Saralegi \& {San Vicente}(2013{\natexlab{a}})}]{Saralegi2013b}
\bibinfo{author}{Saralegi, X.}, \& \bibinfo{author}{{San Vicente}, I.}
  (\bibinfo{year}{2013}{\natexlab{a}}).
\newblock \bibinfo{title}{Elhuyar at tass 2013}.
\newblock In {\it \bibinfo{booktitle}{Proceedings of the TASS 2013 Workshop at
  SEPLN}\/}.
\bibitem[{Saralegi \& {San
  Vicente}(2013{\natexlab{b}})}]{saralegi_tweetnorm_2013}
\bibinfo{author}{Saralegi, X.}, \& \bibinfo{author}{{San Vicente}, I.}
  (\bibinfo{year}{2013}{\natexlab{b}}).
\newblock \bibinfo{title}{Elhuyar at tweetnorm 2013}.
\newblock In {\it \bibinfo{booktitle}{Proceedings of the TweetNorm Workshop at
  SEPLN}\/}.
\bibitem[{Severyn \& Moschitti(2015)}]{severyn2015}
\bibinfo{author}{Severyn, A.}, \& \bibinfo{author}{Moschitti, A.}
  (\bibinfo{year}{2015}).
\newblock \bibinfo{title}{Unitn: Training deep convolutional neural network for
  twitter sentiment classification}.
\newblock In {\it \bibinfo{booktitle}{Proceedings of the 9th International
  Workshop on Semantic Evaluation (SemEval 2015)}\/} (pp.
  \bibinfo{pages}{464--469}).
\bibitem[{Shaikh et~al.(2017)Shaikh, Feldman, Barach \&
  Marzouki}]{shaikh2017tweet}
\bibinfo{author}{Shaikh, S.}, \bibinfo{author}{Feldman, L.~B.},
  \bibinfo{author}{Barach, E.}, \& \bibinfo{author}{Marzouki, Y.}
  (\bibinfo{year}{2017}).
\newblock \bibinfo{title}{Tweet sentiment analysis with pronoun choice reveals
  online community dynamics in response to crisis events}.
\newblock In {\it \bibinfo{booktitle}{Advances in cross-cultural decision
  making}\/} (pp. \bibinfo{pages}{345--356}).
\bibitem[{Shamma et~al.(2011)Shamma, Kennedy \& Churchill}]{shamma2011peaks}
\bibinfo{author}{Shamma, D.~A.}, \bibinfo{author}{Kennedy, L.}, \&
  \bibinfo{author}{Churchill, E.~F.} (\bibinfo{year}{2011}).
\newblock \bibinfo{title}{Peaks and persistence: modeling the shape of
  microblog conversations}.
\newblock In {\it \bibinfo{booktitle}{Proceedings of the ACM 2011 conference on
  Computer supported cooperative work}\/} (pp. \bibinfo{pages}{355--358}).
\newblock \bibinfo{organization}{ACM}.
\bibitem[{Skaperdas \& Grofman(1995)}]{skaperdas_grofman_1995}
\bibinfo{author}{Skaperdas, S.}, \& \bibinfo{author}{Grofman, B.}
  (\bibinfo{year}{1995}).
\newblock \bibinfo{title}{Modeling negative campaigning}.
\newblock {\it \bibinfo{journal}{American Political Science Review}\/},  {\it
  \bibinfo{volume}{89}\/}, \bibinfo{pages}{49–61}.
\bibitem[{Somasundaran et~al.(2009)Somasundaran, Namata, Wiebe \&
  Getoor}]{somasundaran_supervised_2009}
\bibinfo{author}{Somasundaran, S.}, \bibinfo{author}{Namata, G.},
  \bibinfo{author}{Wiebe, J.}, \& \bibinfo{author}{Getoor, L.}
  (\bibinfo{year}{2009}).
\newblock \bibinfo{title}{Supervised and unsupervised methods in employing
  discourse relations for improving opinion polarity classification}.
\newblock In {\it \bibinfo{booktitle}{Proceedings of the 2009 Conference on
  Empirical Methods in Natural Language Processing: Volume 1 -}\/} {EMNLP} '09
  (pp. \bibinfo{pages}{170--179}).
\newblock \bibinfo{address}{Stroudsburg, {PA}, {USA}}.
\bibitem[{Sutton(2009)}]{sutton2009social}
\bibinfo{author}{Sutton, J.~N.} (\bibinfo{year}{2009}).
\newblock \bibinfo{title}{Social media monitoring and the democratic national
  convention: New tasks and emergent processes}.
\newblock {\it \bibinfo{journal}{Journal of Homeland Security and Emergency
  Management}\/},  {\it \bibinfo{volume}{6}\/}.
\bibitem[{Taboada et~al.(2011)Taboada, Brooke, Tofiloski, Voll \&
  Stede}]{taboada2011lexicon}
\bibinfo{author}{Taboada, M.}, \bibinfo{author}{Brooke, J.},
  \bibinfo{author}{Tofiloski, M.}, \bibinfo{author}{Voll, K.}, \&
  \bibinfo{author}{Stede, M.} (\bibinfo{year}{2011}).
\newblock \bibinfo{title}{Lexicon-based methods for sentiment analysis}.
\newblock {\it \bibinfo{journal}{Computational linguistics}\/},  {\it
  \bibinfo{volume}{37}\/}, \bibinfo{pages}{267--307}.
\bibitem[{Thelwall(2017)}]{thelwall2017heart}
\bibinfo{author}{Thelwall, M.} (\bibinfo{year}{2017}).
\newblock \bibinfo{title}{The heart and soul of the web? sentiment strength
  detection in the social web with sentistrength}.
\newblock In {\it \bibinfo{booktitle}{Cyberemotions}\/} (pp.
  \bibinfo{pages}{119--134}).
\bibitem[{Walter(2014)}]{walter2014choosing}
\bibinfo{author}{Walter, A.~S.} (\bibinfo{year}{2014}).
\newblock \bibinfo{title}{Choosing the enemy: Attack behaviour in a multiparty
  system}.
\newblock {\it \bibinfo{journal}{Party Politics}\/},  {\it
  \bibinfo{volume}{20}\/}, \bibinfo{pages}{311--323}.
\bibitem[{Xu et~al.(2012)Xu, Zhu \& Bellmore}]{Xuetal2012sabullying}
\bibinfo{author}{Xu, J.-M.}, \bibinfo{author}{Zhu, X.}, \&
  \bibinfo{author}{Bellmore, A.} (\bibinfo{year}{2012}).
\newblock \bibinfo{title}{Fast learning for sentiment analysis on bullying}.
\newblock In {\it \bibinfo{booktitle}{Proceedings of the First International
  Workshop on Issues of Sentiment Discovery and Opinion Mining}\/} WISDOM '12
  (pp. \bibinfo{pages}{10:1--10:6}).
\newblock \bibinfo{address}{New York, NY, USA}: \bibinfo{publisher}{ACM}.
\newblock \URLprefix \url{http://doi.acm.org/10.1145/2346676.2346686}.
  \DOIprefix\doi{10.1145/2346676.2346686}.
\bibitem[{Yu \& Wang(2015)}]{yu2015world}
\bibinfo{author}{Yu, Y.}, \& \bibinfo{author}{Wang, X.} (\bibinfo{year}{2015}).
\newblock \bibinfo{title}{World cup 2014 in the twitter world: A big data
  analysis of sentiments in us sports fans’ tweets}.
\newblock {\it \bibinfo{journal}{Computers in Human Behavior}\/},  {\it
  \bibinfo{volume}{48}\/}, \bibinfo{pages}{392--400}.
\bibitem[{Zubiaga et~al.(2017)Zubiaga, Voss, Procter, Liakata, Wang \&
  Tsakalidis}]{zubiaga2017towards}
\bibinfo{author}{Zubiaga, A.}, \bibinfo{author}{Voss, A.},
  \bibinfo{author}{Procter, R.}, \bibinfo{author}{Liakata, M.},
  \bibinfo{author}{Wang, B.}, \& \bibinfo{author}{Tsakalidis, A.}
  (\bibinfo{year}{2017}).
\newblock \bibinfo{title}{Towards real-time, country-level location
  classification of worldwide tweets}.
\newblock {\it \bibinfo{journal}{IEEE Transactions on Knowledge and Data
  Engineering}\/},  {\it \bibinfo{volume}{29}\/}, \bibinfo{pages}{2053--2066}.

\end{thebibliography}







\appendix


\newgeometry{margin=1.25cm}
\pagestyle{empty}
\begin{landscape}
	\section{Comparative of commercial Social Media Monitors}
	\label{isv2018:annex:commercial}
	
			\begin{longtable}{p{0.11\textwidth}|p{0.25\textwidth}|p{0.15\textwidth}|p{0.25\textwidth}|p{0.2\textwidth}|p{0.16\textwidth}}
				\hline
				\textbf{Platform} & \textbf{Data Sources} & \textbf{Crawling} & \textbf{Data Processing} & \textbf{Search} & \textbf{Navigation}  \\
				\hhline{======}
				Iconoce & Digital press, blogs, videos, social media (Facebook, Twitter, Linked-in?) & Personalized, subject to agreement  & no & Personalized archive \newline 3 separate search engines (mentions, comment, authors) no lemmatization - no crosslingual. & Graphs (aggregations?), salient terms, salient topics, Influencers, alerts, reports\\			
				\hline
				INNGUMA & Rss multimedia, Deep Web, Twitter, Facebook, Linkedin, possibility to include external documentation manually & Custom filters, not clear if filtering is done at a post-crawling stage & MT, No mention of text processing. No SA & Semantic search (techniques not especified). Index cards and documents. Information is tagged manually. & Reports, content creation, social media management. Multilingual GUI.\\
				\hline
				Meaning Cloud & Digital news, blogs, Twitter, satisfaction surveys (customer provided), phone survey transcriptions, &   & 5 languages (Es, En, Fr, Pt, It). Language identification, Clustering for topic detection. Lemmatization, pos tagging, parsing, NERC, GATE API \newline SA: Ruled-based. Sentiment Lexicons + rules. Irony and subjectivity detection. Entity polarity detected using manually compiled dictionaries. & no & No Dashboard, visualizations or data aggregatios. Excel plugin or API access \\
				\hline
				Snap-trends & Social Media (Twitter, Facebook, Instagram, Google+,...) &  & MT from 80 languages. Propietary linguistic processing. Topic (trends) detection.\newline Propietary sentiment analysis. & Geolocation based search engine, mutiple criteria: social network, search terms, geolocation. Previous search feature. & Agreggations, interactive visualization, temporal trends. \\
				\hline
				Websays & News, Blogs/RSS, Forums, Facebook, Twitter, Google+, LinkedIn, Instagram, Foursquare, Pinterest, Youtube, Vimeo, Reviews (Tripadvisor, Booking,...) & Keyword based, accepts also negative keywords. & Multilingual data processing, no specific data about the coverage \newline SA: AI (ML) + human validation & Multiple search criteria, filter-based. & Graphs, salient terms, trending topics, influencers, sentiment, trends. Alerts and reports. \\
				\hline
				Lynguo & Facebook, Twitter, Instagram, YouTube, online media, blogs and forums. & Keyword based, accepts also negative keywords and accounts. & Es,En. \newline SA: ML + Lexicons + rules. Polarity and emotions. Aspect based SA &  & Customizable dashboard. Several default aggregations and possibility to generate custom visualizations. Alerts and periodical reports \\
				\hline
				Keyhole & Twitter, Instagram, web sources. & Social media and web sources are configured and monitored separately. Keywords, users. & 13 languages. \newline SA: Polarity &  & Influencers, timeline, trends, sentiment, aggregations. \\
				\hline
				Uber-metrics & Blogs, forums, academic/scientific journals, digital press, Instagram, Tumblr, Google+, Facebook, Twitter, YouTube, Vimeo, Flickr, and Foursquare. With Ubermetrics you can even capture comments from YouTube, Facebook, and major online news sources. TV/Radio & Customizable "search agents". Keyword based & 40 languages. Propietary data processing. & Detailed search based on multiple criteria included in visualization dashboard & Dashboard, alerts, reports.  \\
				\hline
			\caption{Comparison of commercial social media monitoring platforms.}
			\label{isv2018:soa:indust:comptable}
			\end{longtable}  
		
\end{landscape}
\pagestyle{plain}
\restoregeometry

\section{Polarity annotation guidelines}
\label{isv2018:annex:guidelines}

We present the guidelines provided to the annotators for marking entity level polarity, including ambiguous cases and the solutions proposed for each of them:

\begin{itemize}
	\item Neutral: There is no clear opinion or sentiment respect to the target party or candidate from the holder. Mentions referring to objective facts fall into this category as well, even if the fact may be considered positive or negative (e.g. \textit{``El PNV consigue grupo en el senado''}\transen{PNV gets its own group in the senate} ).
	\item Positive: The mention includes a positive assessment from the holder with respect to the target (e.g. \textit{``Urkullu ha sido un buen lehendakari.''}\transen{Urkullu has been a good president}). 
	\item Negative: The mention includes a negative assessment from the holder with respect to the target (e.g. \textit{``Urkullu ha sido un lehendakari mediocre.''}\transen{Urkullu has been a mediocre president}). 
	\item ambiguous cases:
	\begin{enumerate}
		\item Subjectivity is not explicit:  \textit{``Cataluña desobedece constantemente la Ley, PNV pide acercamiento de presos, Ribo da los pasos hacia el nacionalismo y Rajoy en SanXenso''}\transen{Catalunya constantly disobeys the law, PNV asks for the rapprochement of prisoners, Ribo makes steps towards nationalism and (meanwhile) Rajoy is in SanXenso.}. Main target in the example is ``Rajoy'' but author expresses a negative opinion towards PNV. Annotators were ask to interpret the implicit subjectivity according to the holder.
		
		\item The holder expresses the opinion of a third party: \textit{``Podemos cree que Urkullu tiene miedo y por eso adelantará las elecciones - EcoDiario.es \textless URL\textgreater''}\transen{Podemos thinks Urkullu is scared and that's why he will call the election early - EcoDiario.es \textless URL\textgreater.}.The mention expresses a negative opinion from Podemos towards Urkullu. Annotators were asked to annotate it as negative towards Urkullu if they could certify that the holder agreed with the opinion from Podemos, or netural otherways.
		
		\item There are two (or more) references to a single target, expressing different polarities: \textit{``PNV tendrá grupo propio en el Senado tras la cesión de cuatro asientos por parte del PP y mantiene su ``no a Rajoy'' ''}\transen{PNV will have its own group in the senate thanks to PP handing over four seats, and they still maintain the "No to Rajoy".}. The following criteria were applied: N+P=NEU, N+NEU=N, P+NEU=P.
		
		\item The polarity of the message and the polarity towards the target are different: \textit{``El tercer precandidato de \#Podemos llama a desalojar al PNV \textless URL\textgreater''}\transen{The number three shortlisted candidate of \#Podemos calls on the people for throwing PNV out \textless URL\textgreater}. In those cases, polarity towards the target should be annotated. In the example, message polarity would be neutral, but polarity towards "PNV" would be negative. Thus message would be marked as negative.
		
		\item Irony/sarcasm: \textit{``¿Las cambiamos por Calle Arnaldo Otegi o Paseo de Juana Chaos? Al fin y al cabo, son hombres de paz... \textless URL\textgreater''}\transen{What if we change the name of the street to Arnaldo Otegi St. or Paseo de Juana Chaos? After all, they are men of peace... \textless URL\textgreater}. Annotators were asked to interpret irony. The previous example would be thus negative towards the target Arnaldo Otegi.
			
		\item The holder is condemning a negative stance against the target: \textit{``@eldiarionorte cada vez se os ve más el plumero. Panfleto anti bildu. Cuando la salud de los zubietarras empeore, vais y se lo contáis.''}\transen{@eldiarionorte it is increasingly clear what you are up to. Anti Bildu pamphlet. When the people in Zubieta lose their health go and tell them.}. Annotators were asked to interpret the intention of the holder. If the notice a clear intention of defending the target the it should be regarded as positive.
		
		\item The target captured is not the main focus of the opinion: \textit{``@CristinaSegui\_ Subió impuestos,no hace nada contra los nacionalistas y les da dinero,no ilegaliza a Bildu y stá implicado en lo de Bárcenas''}\transen{@CristinaSegui\_ He increased taxes, he does nothing against nationalists and gives them money, he does not ban Bildu and he is involved in the Bárcenas affair.}. Annotators were asked to mark the polarity towards the target, regardless of the main focus of the opinion.

	\end{enumerate}
	
\end{itemize} 

\clearpage

\end{document}